\documentclass[11pt,a4paper]{article}
\usepackage[T1]{fontenc}
\usepackage[utf8]{inputenc}
\usepackage{lmodern}
\usepackage[a4paper,margin=25mm,headheight=14pt]{geometry}
\usepackage{amsmath,amssymb,graphicx,booktabs,array,multirow,adjustbox,overpic}
\usepackage{enumitem,placeins,float,algorithm,algpseudocode}
\usepackage[expansion=false]{microtype}
\usepackage{setspace}
\usepackage[font=small,labelfont=bf]{caption}
\usepackage[numbers,super,sort&compress]{natbib}
\usepackage{xcolor}
\PassOptionsToPackage{hyphens}{url}
\usepackage{hyperref}
\hypersetup{colorlinks=true,urlcolor={blue!65!black},linkcolor=black,citecolor=black,pdfauthor={Mrunmay Angaitkar and coauthors}}
\graphicspath{{figures/}}
\newcommand{\base}{\ensuremath{\mathrm{DRR}_{\text{base}}}}
\newcommand{\trans}{\ensuremath{\mathrm{DRR}_{\text{trans}}}}
\newcommand{\mainref}[2]{\ref{#1}}
\begin{document}
\hypersetup{pdftitle={Anatomically decomposed computed tomography projections as scalable supervision for bone suppression in chest radiographs}}
\begin{center}
{\LARGE\bfseries Anatomically decomposed computed tomography projections as scalable supervision for bone suppression in chest radiographs\par}
\vspace{12pt}
{\normalsize Mrunmay Angaitkar\textsuperscript{1,*,$\dagger$}, Piyush Kumar\textsuperscript{1,$\dagger$},
Aarjav Satia\textsuperscript{1,$\ddagger$}, Pranav Rao\textsuperscript{1,$\ddagger$},
Ashish Mittal\textsuperscript{1}, Manoj Tadepalli\textsuperscript{1} and Preetham Putha\textsuperscript{1}\par}
\vspace{8pt}
{\small \textsuperscript{1}Qure.ai, Mumbai, India.\par}
\vspace{6pt}
{\small \textsuperscript{*}Corresponding author.\par}
{\small Email addresses: \href{mailto:mrunmay.angaitkar@qure.ai}{mrunmay.angaitkar@qure.ai} (Mrunmay Angaitkar), \href{mailto:piyush.kumar@qure.ai}{piyush.kumar@qure.ai} (Piyush Kumar), \href{mailto:aarjav.satia@qure.ai}{aarjav.satia@qure.ai} (Aarjav Satia), \href{mailto:pranav.rao@qure.ai}{pranav.rao@qure.ai} (Pranav Rao), \href{mailto:ashish.mittal@qure.ai}{ashish.mittal@qure.ai} (Ashish Mittal), \href{mailto:manoj.tadepalli@qure.ai}{manoj.tadepalli@qure.ai} (Manoj Tadepalli), \href{mailto:preetham.putha@qure.ai}{preetham.putha@qure.ai} (Preetham Putha)\par}
{\small \textsuperscript{$\dagger$}These authors contributed equally and share first authorship.\par}
{\small \textsuperscript{$\ddagger$}These authors contributed equally and share second authorship.\par}
\end{center}
\begin{abstract}
Bone suppression in chest radiographs can improve the visibility of abnormalities obscured by bone, but paired supervision is scarce. We decompose chest computed tomography (CT) volumes into bone, lung and non-lung soft-tissue components and generate aligned digitally reconstructed radiographs. Their weighted sum yields synthetic training pairs on which we train bone- and lung-component suppression models that transfer to real radiographs. Across TBX11K, Node21 and VinDr-CXR with RetinaNet and Faster R-CNN, detectors trained on our suppressed images achieve the highest sensitivity and mean average precision among evaluated suppression methods in all six settings, and combining suppressed with original images improves sensitivity over original-only training in all six. Only our method increases nodule contrast-to-noise ratio (Node21, VinDr-CXR) while retaining clear-lung detail. An unpaired translation extension reduces Fr\'echet inception distance against real radiographs by over 40\% versus the best open-source projection engine, and improves anatomical-structure detection and lung-field agreement with the source CT. Models and inference code: \href{https://huggingface.co/qureaiorg/bone-suppression}{\texttt{huggingface.co/qureaiorg/bone-suppression}}; translated projections: \href{https://huggingface.co/datasets/qureaiorg/ct2xr-projections}{\texttt{huggingface.co/datasets/qureaiorg/ct2xr-projections}}.
\end{abstract}
\noindent\textbf{Keywords:} chest radiography; bone suppression; synthetic supervision; computed tomography; digitally reconstructed radiographs; unpaired domain translation.
\section{Introduction}
\label{sec:introduction}\label{sec:related_drr}
Ribs and clavicles can obscure pulmonary abnormalities on chest radiographs (CXRs). Bone suppression may improve visibility for readers and computer-aided detection~\cite{li2011small,bae2022bone}, provided it retains the underlying findings. Paired supervision is scarce: dual-energy subtraction (DES) requires dedicated acquisition and introduces motion, noise and cross-contamination artefacts~\cite{loog2006filter,chen2014separation}. Adversarial and diffusion methods learn from these pairs~\cite{rajaraman2021bs,liu2020egan,bsldm,gllcm}. Alternatively, forward projection simulates X-ray paths through a three-dimensional chest computed tomography (CT) volume to form a two-dimensional synthetic radiograph, termed a digitally reconstructed radiograph (DRR). Projecting separated bone and soft tissue through the same geometry yields aligned training pairs, as established by prior CT-supervised suppression studies~\cite{gozes2020bone,ren2021ctfeatures}.

Standard engines such as DeepDRR, DiffDRR, Plastimatch and TIGRE~\cite{deepdrr,diffdrr,plastimatch,tigre} address projection rather than complete suppression-pair construction. Engines accepting tissue labels or material-assigned meshes, including gVirtualXray~\cite{gvxr}, still require anatomical decomposition. Separating bone in CT volumes via naive thresholding can retain isolated noisy voxels and mislabel dense non-osseous tissue as bone. When projected, these segmentation errors can produce spurious structures resembling lung opacities, nodules or calcifications in the bone targets, corrupting the training signal. Bone removal also leaves voids in the remaining soft-tissue volume, producing rib-shaped imprints in its projection. A single conversion from CT Hounsfield units (HU) to X-ray attenuation also limits contrast diversity because attenuation depends on photon energy and tissue composition~\cite{watanabe1999lac}.

\begin{figure}[H]
\centering
\includegraphics[width=160.0mm]{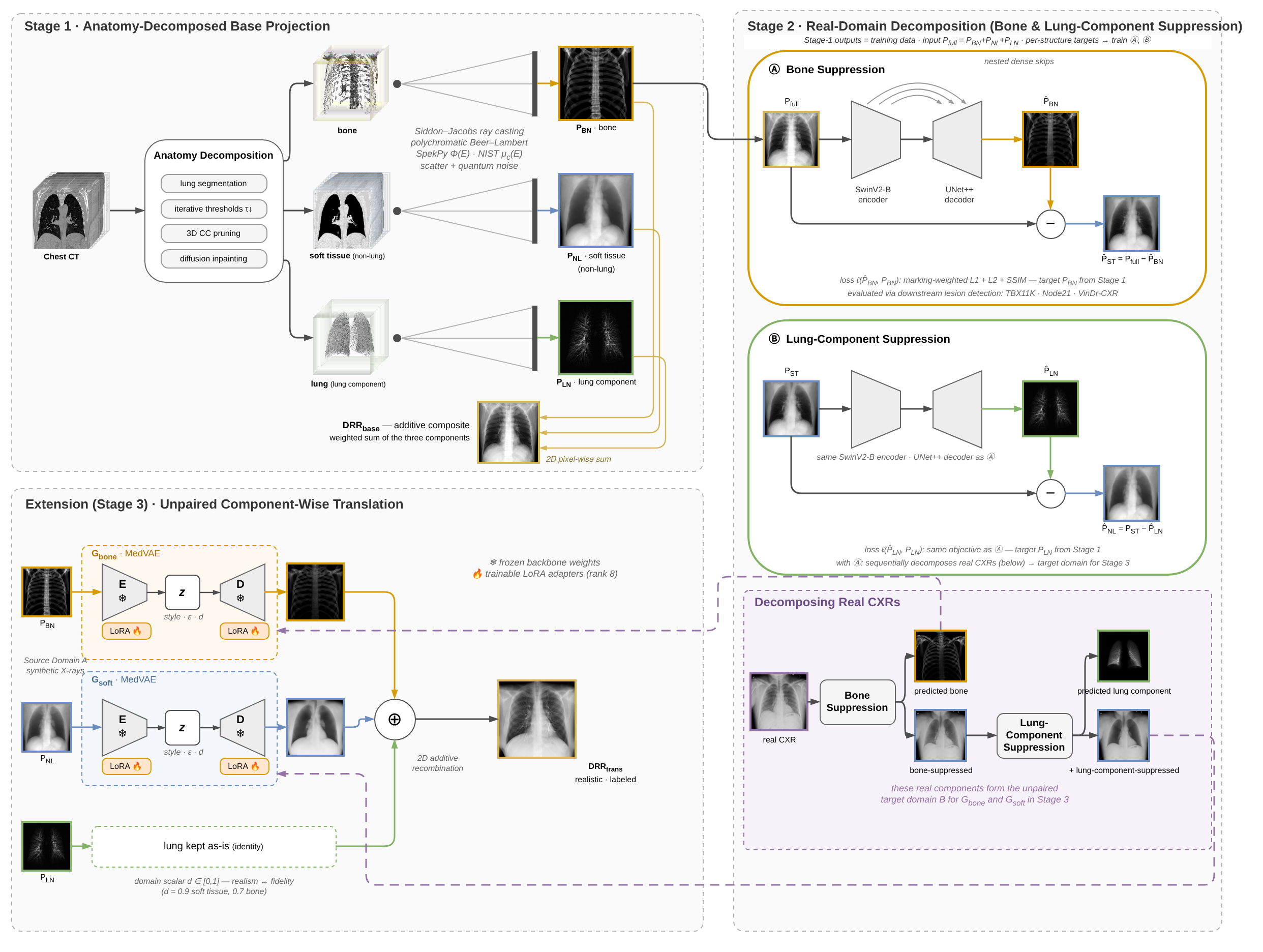}
\caption{\textbf{Framework for anatomical supervision and its translation extension.} Overview of the two-stage framework and its translation extension. Stage~1: Anatomy decomposition and per-tissue polychromatic projection yield pixel-registered per-structure projections that sum to the full radiograph. Stage~2: Bone- and lung-component suppression models trained on Stage 1 DRRs; applied to real radiographs, the suppression models decompose them into components. Stage~3: Component-wise unpaired translation, whose target domain is the real components produced by Stage~2.}
\label{fig:pipeline}
\end{figure}

Our framework (Fig.~\ref{fig:pipeline}) combines iterative bone segmentation, removal of small disconnected regions from the bone mask, filling of the spaces left by bone removal (void inpainting), and lung segmentation to anatomically decompose a CT volume into bone, lung and non-lung soft-tissue components. Stage~1 renders each tissue using attenuation over an X-ray energy spectrum (polychromatic rendering). The aligned bone, lung and non-lung soft-tissue images sum to the base synthetic radiograph, \base{}. Stage~2 trains two feedforward models: bone prediction and subtraction first leave a soft-tissue image; lung-component prediction and subtraction then separate the pulmonary vessels, parenchyma and intrapulmonary findings from non-lung soft tissue. Thus, the second model performs lung-component suppression, which removes the projected contribution of lung tissue, including pulmonary vessels. Training targets are synthetic (real CXRs only provide intensity-distribution examples for histogram-matching augmentation). We evaluate suppression on downstream detection for TBX11K (tuberculosis lesions), Node21 (pulmonary nodules) and VinDr-CXR (14 chest-finding classes), alongside lesion conspicuity and bone-overlap analyses. These experiments evaluate whether a bone-suppression model trained on synthetic projections improves lesion detection when applied to real radiographs, with detectors trained on real-image inputs.

The component images extracted from real radiographs provide appearance examples for Stage~3, which learns from synthetic and real images without requiring matched pairs (unpaired translation). Such mappings can be constrained by cycle consistency, contrastive learning or adversarial diffusion~\cite{cyclegan,cut,syndiff}; we adapt shortest-path regularisation~\cite{santa} to change the appearance of bone and non-lung soft tissue while constraining anatomical changes. Recombining these translated images with the unchanged lung image gives the translated DRR, \trans{}. We assess the translated DRRs' similarity to real-radiograph distributions and their anatomical agreement with the source CTs. Prior work on synthetic radiographs and lesion augmentation~\cite{schultheiss2021synthetic,shen2023synthesis} motivates future evaluation of these projections to augment training data for diagnostic models.

\section{Results}
\label{sec:results}
\begin{figure}[!htb]
\centering
\vspace{6mm}
\begin{overpic}[width=160.0mm]{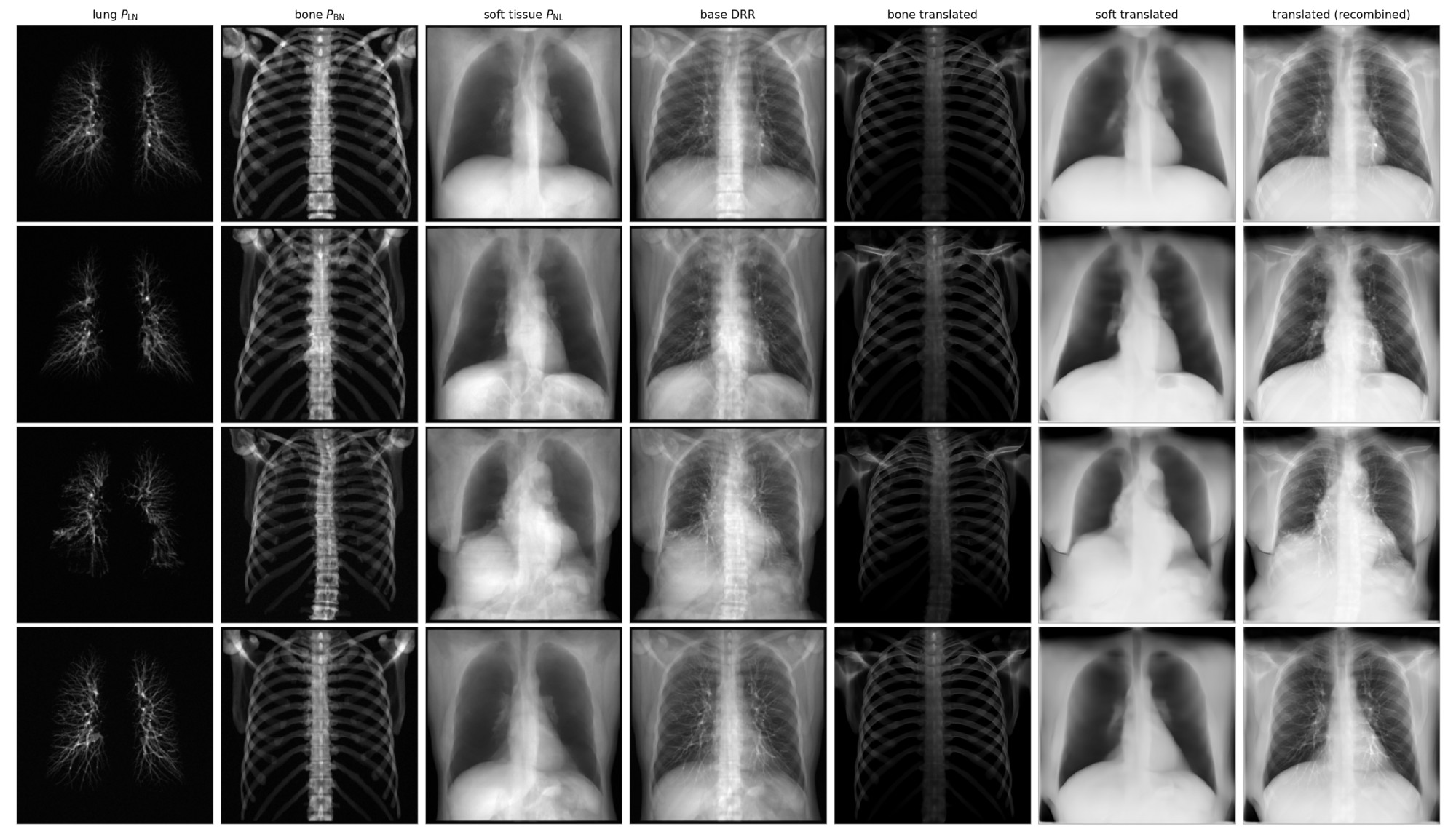}
\put(1.00,58.35){\sffamily\bfseries\small a}
\put(15.29,58.35){\sffamily\bfseries\small b}
\put(29.57,58.35){\sffamily\bfseries\small c}
\put(43.86,58.35){\sffamily\bfseries\small d}
\put(58.14,58.35){\sffamily\bfseries\small e}
\put(72.43,58.35){\sffamily\bfseries\small f}
\put(86.71,58.35){\sffamily\bfseries\small g}
\end{overpic}
\caption{\textbf{Anatomical components and reconstructed radiographs.} Panels a--g correspond to the seven columns in the order described below. From CT to synthetic and translated radiograph, stage by stage (one CT per row). Left to right: the three component projections (lung/vascular $P_{\mathrm{LN}}$, bone $P_{\mathrm{BN}}$, non-lung soft tissue $P_{\mathrm{NL}}$); the base DRR; the translated bone and soft-tissue components; and the recombined translated radiograph \trans{}. The lung component is left untouched by translation.}
\label{fig:stages}
\end{figure}

The framework generates aligned CT projections and learns component suppression; its Stage~3 extension translates the component projections towards real-radiograph appearance (Figs.~\ref{fig:pipeline} and~\ref{fig:stages}). We evaluate suppression through lesion detection on three real-radiograph datasets---TBX11K, Node21 and VinDr-CXR---using two detectors and three training seeds per configuration, and evaluate generated projections for distributional realism and anatomical fidelity.

\subsection{Bone suppression and lesion detection}
\label{sec:bs_results}\label{sec:bs_sota}
\begin{figure}[!htb]
\centering
\vspace{6mm}
\includegraphics[width=160mm]{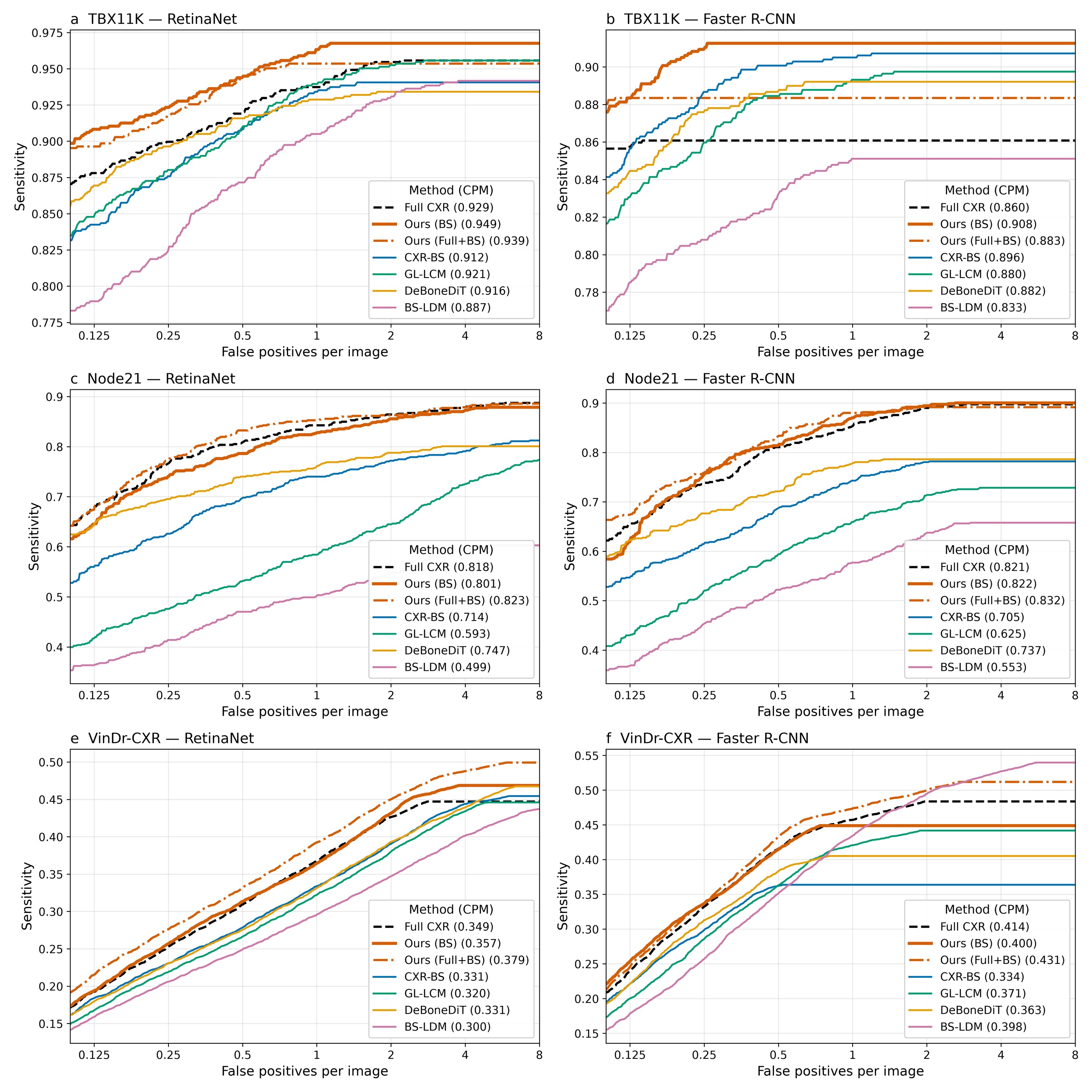}
\caption{\textbf{Lesion detection across three datasets and two detectors.} Panels a--f show TBX11K, Node21 and VinDr-CXR (rows), with RetinaNet and Faster R-CNN (columns). FROC curves show three-seed mean sensitivity for original-image inputs, our suppressed-only and combined inputs, and the evaluated suppressed-only comparison methods. Legends report the mean per-seed CPM from Table~\ref{tab:detection}, calculated over seven operating points from $0.125$ to $8$ false positives per image.}
\label{fig:froc}
\end{figure}

Our bone-suppressed-only inputs achieve the highest mean CPM and mAP@50 among the evaluated suppression systems across all six dataset--detector settings: TBX11K tuberculosis lesions, Node21 nodules and VinDr-CXR's 14 chest-finding classes, each evaluated with RetinaNet and Faster R-CNN (Table~\ref{tab:detection}; Fig.~\ref{fig:froc}). Results use three training seeds on fixed partitions for every dataset. RetinaNet~\cite{retinanet} directly predicts lesion boxes and classes; Faster R-CNN~\cite{fasterrcnn} first proposes candidate regions and then classifies and refines them. For each architecture, we train separate detectors on the original radiograph ($full$), the bone-suppressed image ($bs$), or both images stacked as two channels ($full_{bs}$), holding the training protocol fixed across input configurations within each dataset. Detection is measured by mean sensitivity across seven false-positive rates (FROC CPM) and mean average precision at a bounding-box intersection-over-union threshold of 0.5 (mAP@50); higher values are better.

The benefit relative to original images depends on the dataset and input configuration. On TBX11K, our suppressed-only input increases CPM by $0.020$ and $0.048$ for RetinaNet and Faster R-CNN, respectively, whereas Faster R-CNN mAP@50 is highest with combined input. On Node21, suppressed-only CPM changes by $-0.017/ +0.001$ (RetinaNet/Faster R-CNN), while combined input gains $0.005/ +0.011$. On VinDr-CXR, suppressed-only CPM changes by $+0.008/ -0.014$, while combined input gains $0.030/ +0.017$ and increases mAP@50 from $0.097$ to $0.102$ and from $0.113$ to $0.120$. Our combined input improves mean CPM over original images in all six settings. It also improves mean mAP@50 in five of six settings (Node21 RetinaNet changes from $0.629$ to $0.619$). Thus, our method leads the suppressed-only comparison across all three datasets, while combining it with the original image extends the detection benefit across both architectures.

\begin{figure}[!htb]
\centering
\vspace{6mm}
\begin{overpic}[width=160.0mm]{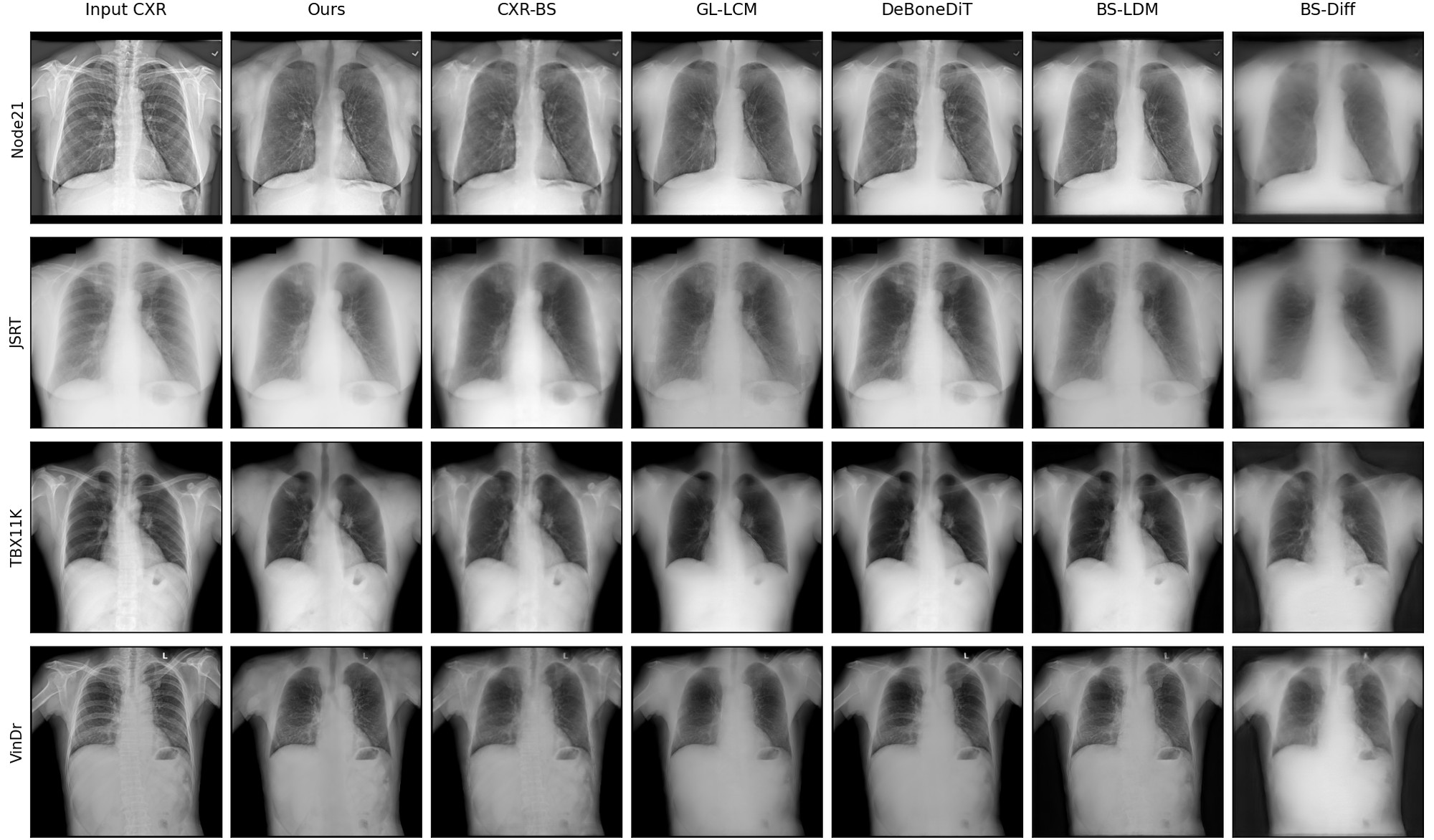}
\put(1.00,59.80){\sffamily\bfseries\small a}
\put(15.29,59.80){\sffamily\bfseries\small b}
\put(29.57,59.80){\sffamily\bfseries\small c}
\put(43.86,59.80){\sffamily\bfseries\small d}
\put(58.14,59.80){\sffamily\bfseries\small e}
\put(72.43,59.80){\sffamily\bfseries\small f}
\put(86.71,59.80){\sffamily\bfseries\small g}
\end{overpic}
\caption{\textbf{Comparison of suppression outputs on real radiographs.} Panels a--g show input CXR, ours, CXR-BS, GL-LCM, DeBoneDiT, BS-LDM and BS-Diff, respectively. Bone suppression on real radiographs, across all four evaluation sources (rows) and every method (columns). Ours removes ribs and clavicles while the vascular tree, diaphragm, and mediastinal borders stay in place; the diffusion baselines flatten the parenchyma and CXR-BS leaves clavicles and rib margins standing.}
\label{fig:bs_sota}
\end{figure}

We compare with CXR-BS~\cite{rajaraman2021bs}, GL-LCM~\cite{gllcm}, BS-LDM~\cite{bsldm} and DeBoneDiT (\url{https://huggingface.co/diaoquesang/DeBoneDiT}) on all three detection datasets. The three diffusion-based systems share a private DES training dataset and method lineage. None of the other methods improves suppressed-only RetinaNet CPM over the original-image baseline on any of the three datasets. Several improve TBX11K Faster R-CNN CPM, but by less than ours; on Node21 and VinDr, ours likewise has the highest suppressed-only CPM among the evaluated methods. Combined input can recover performance through the original-image channel, with ours achieving the highest mean CPM and mAP@50 among suppression methods for both VinDr detectors. BS-Diff~\cite{bsdiff}, the predecessor to the diffusion-based systems, was weakest on TBX11K and its suppressed-only detector failed to train on Node21 (Supplementary Table~\ref{supp-tab:bsdiff}). Figure~\ref{fig:bs_sota} illustrates retained vascular, diaphragmatic and mediastinal detail in our outputs, alongside parenchymal flattening or residual bone margins in the comparison methods.

\subsection{Exploratory bone-overlap and preservation analyses}
\label{sec:analysis}\label{sec:prevalence}
We examine whether detection benefits from suppression are greater for bone-overlapped lesions than for clearer lesions. This exploratory analysis characterises where suppression helps, separately from the overall comparison between suppression methods. Lesions are classified as overlapped when at least $15\%$ of their annotated bounding box intersects our predicted bone mask, and clearer otherwise. Recall is measured at two false positives per image. Within Node21, recall gains favour overlapped nodules for both detectors and input configurations across three training seeds (Table~\ref{tab:subgroup}; Supplementary Fig.~\ref{supp-fig:subgroup}). On VinDr, combined input improves both groups: overlapped/clearer recall gains are $+0.021/ +0.024$ for RetinaNet and $+0.046/ +0.011$ for Faster R-CNN. TBX11K RetinaNet recall is near ceiling; Faster R-CNN gains favour overlapped lesions with combined input and clearer lesions with suppressed-only input. Across methods, our suppressed-only input has the highest mean recall in both groups for five of the six dataset--detector settings (Supplementary Tables~\ref{supp-tab:overlap_tbx11k}--\ref{supp-tab:overlap_vindr}). For VinDr Faster R-CNN, BS-LDM has the highest suppressed-only recall in both groups ($0.405$ overlapped and $0.511$ clearer, versus $0.338$ and $0.469$ for ours), while ours has the highest mean recall in both groups with combined input.

Across VinDr-CXR classes, bone-overlap prevalence ranges from $36.0\%$ for pneumothorax to $1.6\%$ for cardiomegaly (Supplementary Table~\ref{supp-tab:class_prevalence}). Correlating prevalence with seed-averaged class-level CPM changes gives positive associations for our method with both detectors: Spearman $\rho=0.398$ ($p=0.159$) for RetinaNet and $\rho=0.332$ ($p=0.246$) for Faster R-CNN (Table~\ref{tab:perclass}). These results describe a tendency towards larger gains in more overlapped classes; none of the 20 exploratory coefficients meets the nominal $p<0.05$ threshold. Each correlation uses the 14 classes as observations, with CPM changes averaged across three matched training seeds; $p$ values are unadjusted for multiple testing.

Between datasets, overlap is most prevalent on TBX11K (nearly half of lesions), which shows the clearest detection gain; over half of VinDr findings have no bone contact (Table~\ref{tab:prevalence}). The larger gain on TBX11K is consistent with a greater benefit when bone obscures lesions.

\subsubsection{Lesion conspicuity}
\label{sec:conspicuity}
\begin{figure}[!htb]
\centering
\vspace{6mm}
\begin{overpic}[width=160.0mm]{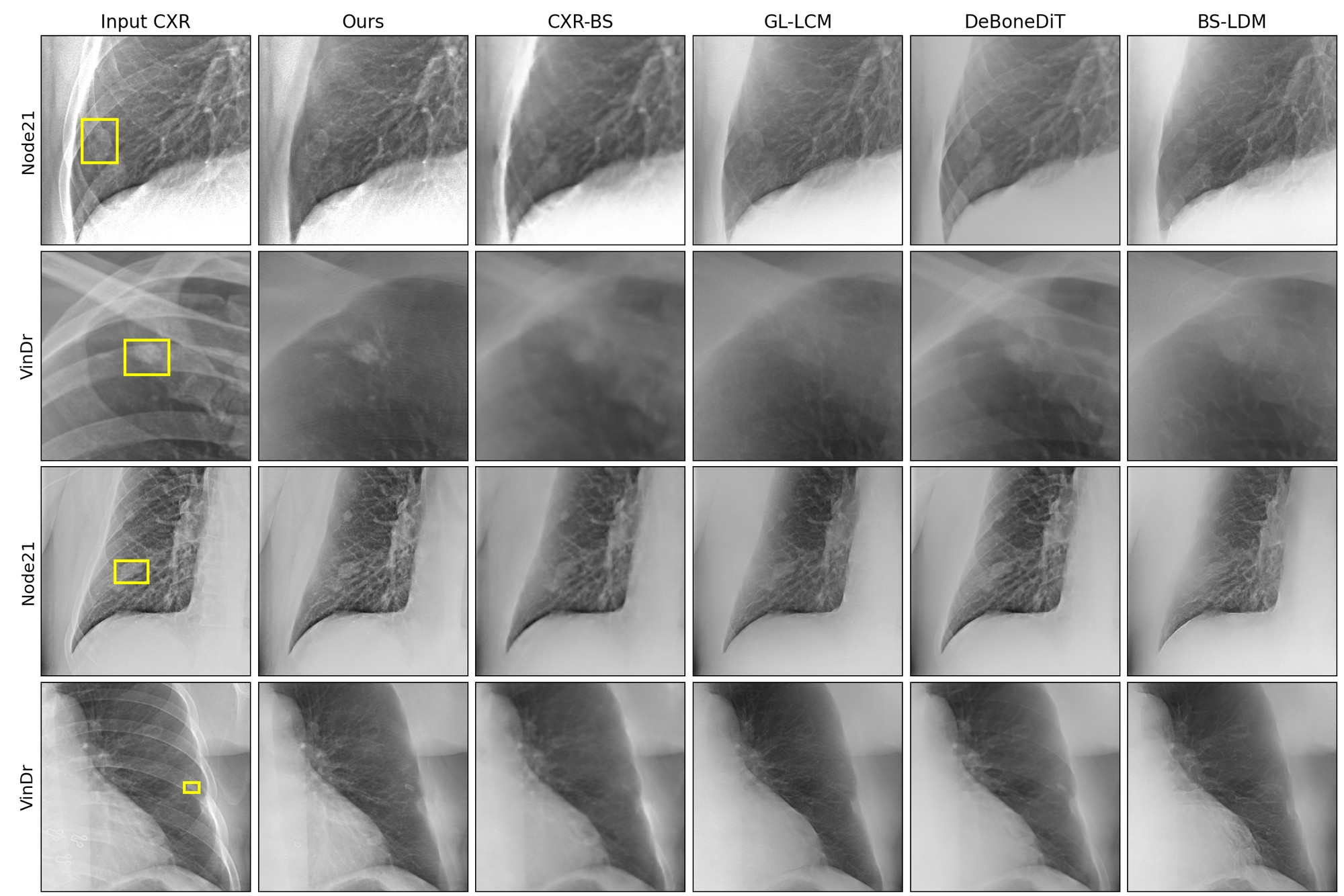}
\put(1.00,67.90){\sffamily\bfseries\small a}
\put(17.67,67.90){\sffamily\bfseries\small b}
\put(34.33,67.90){\sffamily\bfseries\small c}
\put(51.00,67.90){\sffamily\bfseries\small d}
\put(67.67,67.90){\sffamily\bfseries\small e}
\put(84.33,67.90){\sffamily\bfseries\small f}
\end{overpic}
\caption{\textbf{Lesion conspicuity and clear-lung detail.} Panels a--f show input CXR, ours, CXR-BS, GL-LCM, DeBoneDiT and BS-LDM, respectively. Lesion conspicuity (top two rows) and clear-lung detail retention (bottom two rows); the yellow box on the input marks the lesion, and all panels in a row show the same crop. \textbf{Rows 1--2:} bone-overlapped nodules. Ours removes the overlying rib while the nodule stays sharp against the parenchyma; the DES-trained diffusion methods flatten it into the background. \textbf{Rows 3--4:} wider clear-lung views. Fine lung markings appear retained in these predict-and-subtract examples; CXR-BS's $256 \times 256$ output, shown upscaled, cannot resolve them, and the diffusion baselines regenerate soft tissue with markings that no longer follow the input.}
\label{fig:bs_quality}
\end{figure}

To assess lesion conspicuity, we measure contrast-to-noise ratio (CNR): the intensity difference between a lesion and its surrounding background, divided by background intensity variation. Positive suppressed-minus-original change ($\Delta$CNR) indicates greater local contrast (Table~\ref{tab:cnr}; Fig.~\ref{fig:bs_quality}). The analysis includes lesions whose annotated box lies at least $25\%$ inside the rib and clavicle region, selected identically for every method (Methods). For Node21 nodules and VinDr nodule/mass, only ours has a $95\%$ interval entirely above zero: $+0.067\,[+0.025,+0.109]$ (Node21) and $+0.049\,[+0.022,+0.076]$ (VinDr). CXR-BS intervals span zero; the three DES-trained diffusion methods' intervals lie below it. On the Japanese Society of Radiological Technology (JSRT) chest-radiograph dataset~\cite{jsrt}, whose nodules are larger, higher-contrast and almost universally rib-overlapped, our interval includes zero ($+0.008\,[-0.022,+0.039]$), while all baselines' intervals lie below it, including CXR-BS ($-0.050\,[-0.083,-0.018]$). All methods improve large TB lesions (confidence intervals above zero; CXR-BS $+0.055$, ours $+0.047$).

\subsubsection{Bone removal and detail retention}
\label{sec:antiblur}
\begin{figure}[!htb]
\centering
\vspace{6mm}
\begin{overpic}[width=160.0mm]{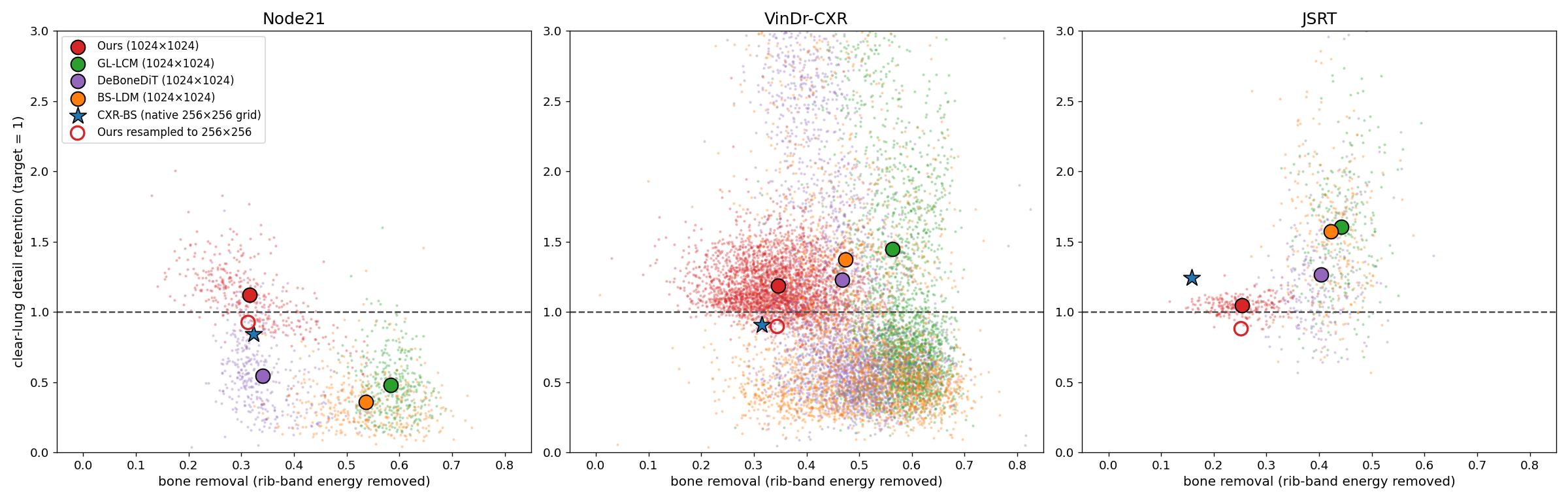}
\put(1.00,33.20){\sffamily\bfseries\small a}
\put(34.33,33.20){\sffamily\bfseries\small b}
\put(67.67,33.20){\sffamily\bfseries\small c}
\end{overpic}
\caption{\textbf{Bone removal and detail retention.} Panels a--c show Node21, VinDr-CXR and JSRT, respectively. Bone removal (rib-band energy removed inside the bone mask) against clear-lung detail retention (target $=1$, dashed) at $1024 \times 1024$ for the $1024 \times 1024$-native methods, on Node21, VinDr, and JSRT; small points are images, large points the per-method means. CXR-BS ($256 \times 256$-native) is shown as a star at its native-grid value, with our resampled $256 \times 256$ value as an open circle. Faithful suppression lies to the right \emph{and} on the dashed line.}
\label{fig:antiblur_scatter}
\end{figure}

Removing bone by blurring the image could also erase useful lung detail. We therefore measure bone removal as the fraction of rib-scale image variation removed within a bone mask, and detail retention as the output-to-input ratio of Laplacian variance (a sharpness measure) in lung regions without bone. Retention near one indicates similar sharpness; lower values suggest blur and higher values suggest added high-frequency detail. Diffusion systems remove more rib-band energy than ours ($0.34$--$0.58$ versus $0.25$--$0.35$), but retain only $0.36$--$0.55$ of Node21 clear-lung Laplacian variance; on VinDr/JSRT their retention exceeds unity with high variability ($1.23$--$1.61$, standard deviation (SD) $0.4$--$1.5$). Our retention is $1.12$, $1.19$ and $1.05$, with SD $0.06$--$0.24$ (Table~\ref{tab:antiblur}; Fig.~\ref{fig:antiblur_scatter}). Our retention values remain closer to unity and less variable than those of the native-resolution diffusion methods across these datasets. Values above unity indicate increased high-frequency variation, which can arise from sharpening or generated texture.

At native $256^2$ resolution, CXR-BS is comparable to ours on Node21/VinDr (removal $0.32/0.31$ versus $0.31/0.34$; retention $0.84/0.91$ versus $0.92/0.90$). On JSRT it removes less ($0.16$ versus $0.25$) and increases high-frequency energy (retention $1.24$), consistent with added texture. The lesion-conspicuity and detection comparisons distinguish our method from CXR-BS despite their comparable removal--retention balance on Node21 and VinDr. When all methods are compared on a common $256^2$ grid, the diffusion methods again remove more rib-band energy than ours, whereas their retention falls below ours on all three datasets ($0.33$--$0.76$ versus $0.88$--$0.92$), because resampling discards the fine-scale variation that raises their native-resolution values (Supplementary Table~\ref{supp-tab:supp_antiblur256}).

\subsection{Real-radiograph component decomposition}
\begin{figure}[!htb]
\centering
\vspace{6mm}
\begin{overpic}[width=152.4mm]{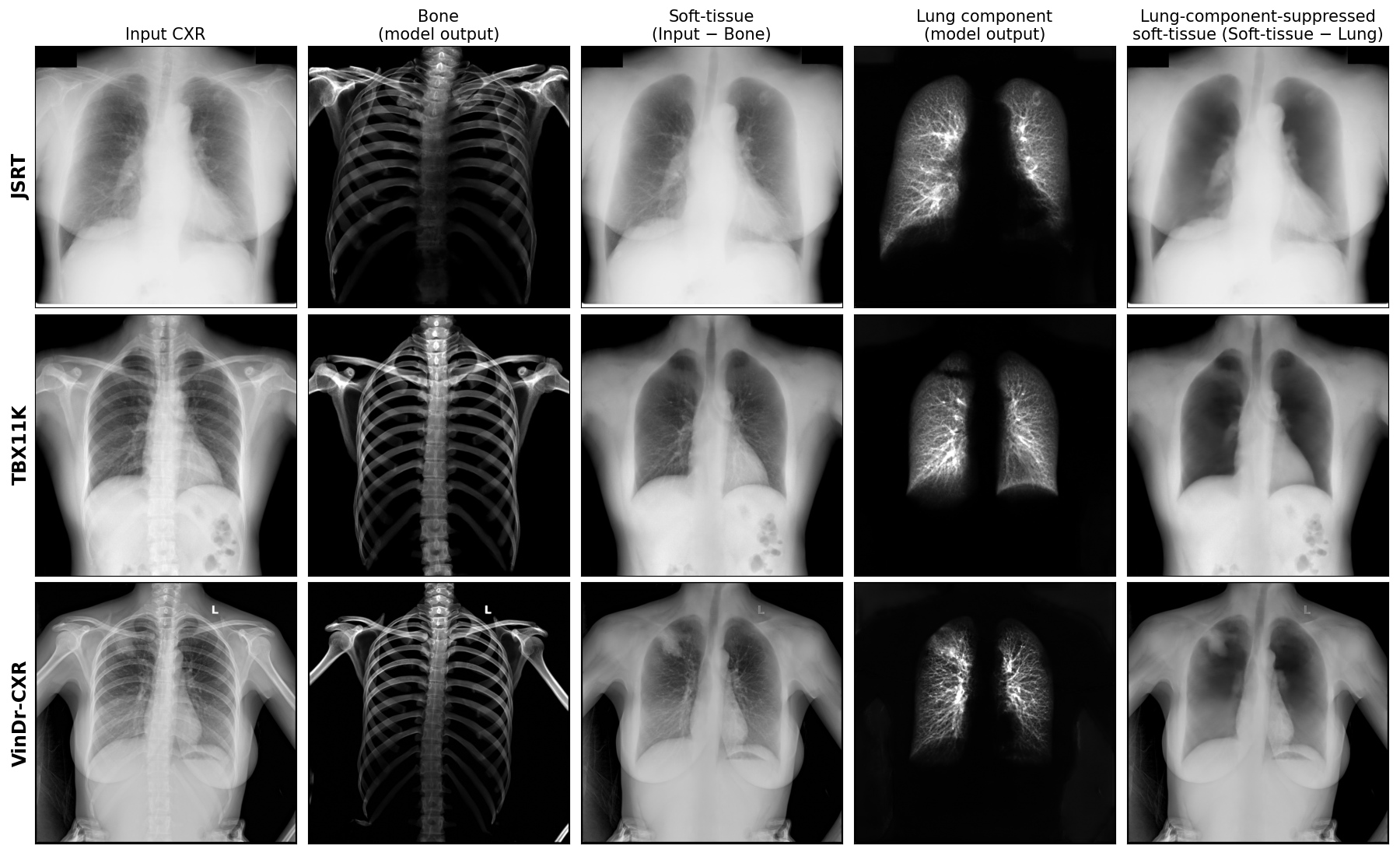}
\put(1.00,62.31){\sffamily\bfseries\small a}
\put(21.00,62.31){\sffamily\bfseries\small b}
\put(41.00,62.31){\sffamily\bfseries\small c}
\put(61.00,62.31){\sffamily\bfseries\small d}
\put(81.00,62.31){\sffamily\bfseries\small e}
\end{overpic}
\caption{\textbf{Component decomposition of real radiographs.} Panels a--e correspond to the five columns in the order described below. Real radiographs decomposed into their components by the bone- and lung-component suppression models applied in sequence (one real dataset per row). Left to right: input CXR; predicted bone image; soft tissue by subtraction (input $-$ bone); predicted lung-component image; lung-component-suppressed soft tissue by a second subtraction (the real-domain counterparts of the Stage-1 components $P_{\mathrm{BN}}$, $P_{\mathrm{ST}}$, $P_{\mathrm{LN}}$, $P_{\mathrm{NL}}$). The bone and lung-component-suppressed soft-tissue columns supply the real target domain for the component-wise translation (Sec.~\ref{sec:translation}).}
\label{fig:vessel}
\end{figure}

The two Stage-2 models decompose a real radiograph in sequence (Fig.~\ref{fig:vessel}). The first predicts bone; subtracting it leaves soft tissue containing both lung and non-lung anatomy. The second predicts the lung component, including pulmonary vessels, parenchyma and intrapulmonary findings, from this soft-tissue image. A second subtraction leaves non-lung soft tissue. The predicted bone, predicted lung component and remaining non-lung soft tissue therefore sum to the original image by construction. Examples span JSRT, TBX11K and VinDr-CXR without dataset-specific tuning. These model-derived components supply the translation targets.

\subsection{Projection realism and anatomical fidelity}
\label{sec:quality}
\begin{figure}[!htb]
\centering
\vspace{6mm}
\begin{overpic}[width=160.0mm]{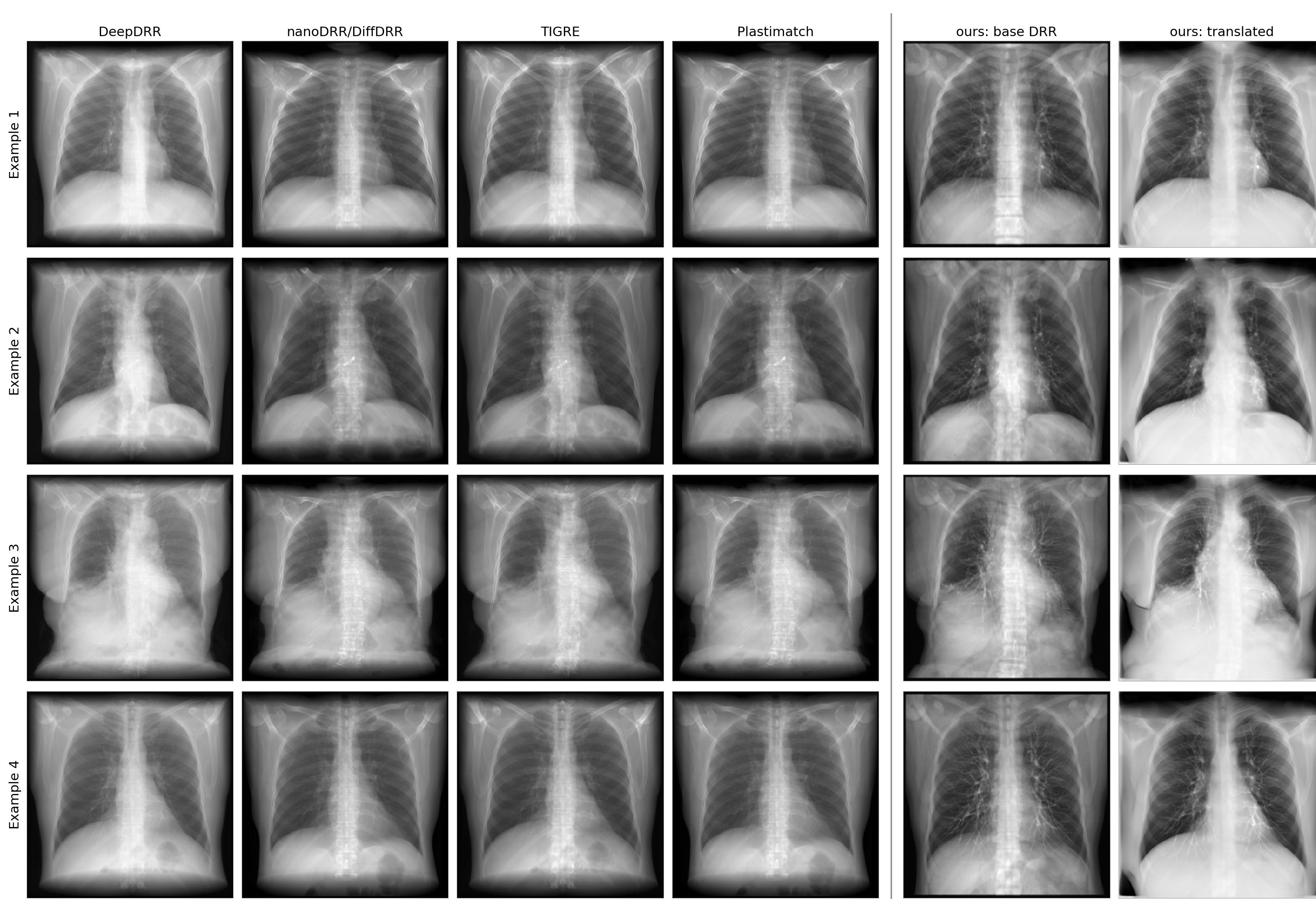}
\put(1.00,70.15){\sffamily\bfseries\small a}
\put(17.67,70.15){\sffamily\bfseries\small b}
\put(34.33,70.15){\sffamily\bfseries\small c}
\put(51.00,70.15){\sffamily\bfseries\small d}
\put(67.67,70.15){\sffamily\bfseries\small e}
\put(84.33,70.15){\sffamily\bfseries\small f}
\end{overpic}
\caption{\textbf{Comparison of projection methods.} Panels a--f show DeepDRR, nanoDRR/DiffDRR, TIGRE, Plastimatch, our base DRR and our translated DRR, respectively. Four source CTs (one per row, the same CTs as Fig.~\ref{fig:stages}), each rendered by the four open-source DRR engines (left: DeepDRR, nanoDRR/DiffDRR, TIGRE, Plastimatch) and by our pipeline (right: base DRR, translated). The line-integral projectors (nanoDRR/DiffDRR, TIGRE, Plastimatch) produce flat, low-contrast images; DeepDRR's poly-energetic model darkens the lungs but stays visibly synthetic. Our \base{} (the additive sum of the per-tissue renders) already shows real-radiograph-like inter-tissue contrast and vascular detail, and translation moves texture further toward real CXR. All panels are shown as scored.}
\label{fig:sota_qual}
\end{figure}

Component-wise translation achieves the lowest FID and KID against both real-radiograph reference pools among the evaluated generators (Table~\ref{tab:fid}). We compare two synthetic outputs: \base{}, assembled directly from the CT-derived component projections, and \trans{}, assembled after translating bone and non-lung soft tissue towards real-CXR appearance while leaving lung unchanged. We render chest CTs from the public CT-RATE dataset~\cite{ctrate} with our framework and four released DRR engines: DeepDRR, Plastimatch, nanoDRR (an optimised DiffDRR implementation) and TIGRE. Real radiographs from VinDr-CXR and CheXpert~\cite{chexpert} provide the reference image distributions (Methods; Fig.~\ref{fig:sota_qual}). Fr\'echet inception distance (FID)~\cite{fid} and kernel inception distance (KID)~\cite{kid} measure differences between synthetic and real image distributions in a CXR-pretrained feature space; lower values indicate closer agreement. KID values below are scaled by $10^3$, as in Table~\ref{tab:fid}. The base FID is within the baseline range against VinDr ($14.8$ versus $14.5$--$16.5$) and lower against CheXpert ($23.6$ versus $27.9$--$29.7$; KID $53$ versus $60$--$65$). Translation has the lowest FID/KID against both references: VinDr $8.2/13.9$ versus best-baseline $14.5/29.6$ (Plastimatch), and CheXpert $16.3/32.0$ versus $27.9/60.4$ (DeepDRR; Table~\ref{tab:fid}).

We also compare local lung-texture entropy, which measures variation in neighbourhood intensity distributions; radially averaged power spectra, which describe image variation at different spatial frequencies; and Laplacian-variance sharpness~\cite{pechpacheco2000}, which measures fine-scale intensity variation. Entropy and spectral distances quantify agreement with real-CXR texture and frequency content, respectively (Methods, Sec.~\ref{sec:projection_metrics}). The base lung-texture entropy distance is lower than all baselines ($0.40$ versus $0.75$--$0.95$), whereas its spectral distance is marginally above the baseline range ($19.7$ versus $15.8$--$19.5$); whole-image sharpness lies within the baseline range ($16.9$ versus $12.7$--$28.1$) and lung-field sharpness is higher ($18.1$ versus $6$--$9$). Translation reduces spectral distance to $2.27$ ($6.9\times$ closer than the best baseline) and entropy distance to $0.31$ ($2.4\times$ closer); whole-image/lung-field sharpness reaches $31.0/23.6$, versus baseline maxima $28.1/9.2$ (Table~\ref{tab:detail}). Real VinDr lung-field sharpness is $291$, indicating a remaining difference in fine-scale image variation despite the gains over the other generators.

\label{sec:fidelity_sub}
Three complementary checks assess fidelity (Table~\ref{tab:fidelity}). First, a classifier trained on real CheXpert CXRs and held fixed predicts four findings from each synthetic image; predictions are scored against labels from the source CT. AUROC-4 averages the areas under the receiver operating characteristic curves for pleural effusion, atelectasis, cardiomegaly and consolidation. This score is $0.768$ for the base, versus $0.772$--$0.782$ for the line-integral projection engines (Plastimatch, nanoDRR and TIGRE) and $0.800$ for DeepDRR. Translation yields $0.774$, combining substantially lower FID with a similar level of finding discrimination to the base projections. Base consolidation AUROC ties DeepDRR at $0.773$; atelectasis is hardest for all generators, with DeepDRR leading effusion and atelectasis.

Second, an independent CXR anatomy segmentation model measures how many of 14 anatomical structures are detected per image. It detects an average of $12.95$ structures in the base projections (baseline range $12.5$--$13.0$; DeepDRR $13.02$), increasing to $13.42$ after translation. The distance between synthetic and real distributions of cardiothoracic ratio (heart width relative to thoracic width) decreases from $0.078$ to $0.025$, versus $0.056$--$0.088$ for baselines ($2.2$--$3.5\times$ closer). Third, the intersection-over-union (IoU) between the lung mask estimated from each synthetic image and the rectangular region projected from its source CT lung box increases from $0.449$ to $0.457$, above all baselines ($0.41$--$0.42$; DeepDRR $0.408$). These results indicate improved lung-field geometric agreement with the source CT.

\subsection{Ablation studies}
\label{sec:ablation}\label{sec:abl_proj}
\begin{figure}[!htb]
\centering
\vspace{6mm}
\begin{overpic}[width=160.0mm]{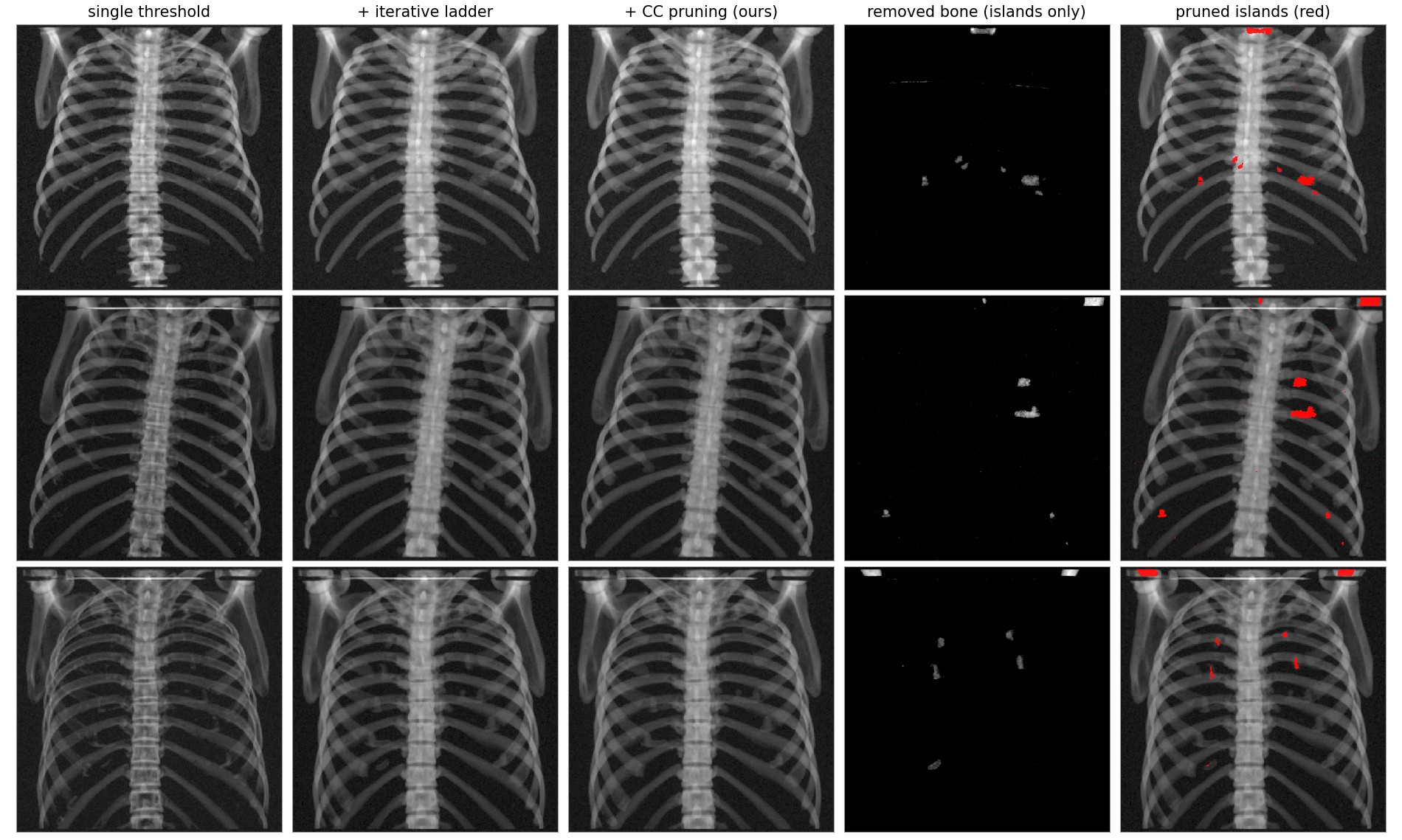}
\put(1.00,61.15){\sffamily\bfseries\small a}
\put(21.00,61.15){\sffamily\bfseries\small b}
\put(41.00,61.15){\sffamily\bfseries\small c}
\put(61.00,61.15){\sffamily\bfseries\small d}
\put(81.00,61.15){\sffamily\bfseries\small e}
\end{overpic}
\caption{\textbf{Effect of bone-mask construction.} Panels a--e correspond to the five columns in the order described below. Effect of bone-mask construction on the bone projection (one CT per row). Left to right: a single fixed threshold; $+$ iterative multi-threshold ladder; $+$ 3D connected-component pruning (ours); the bone removed by pruning alone; and those islands marked in red on the projection; without pruning they project as isolated opacities. Pruning collapses the mask from hundreds of 3D connected components to a handful (e.g.\ $149\rightarrow4$ in the first row).}
\label{fig:abl_boneseg}
\end{figure}

On a fixed subset of $5{,}000$ CT-RATE volumes, we first test the successive steps of our CT bone-segmentation algorithm: a fixed HU threshold, iterative threshold relaxation with neighbourhood filtering, and removal of small disconnected regions (pruning). Here, a connected component is a spatially connected group of bone-labelled voxels in the three-dimensional mask, distinct from the anatomical bone/lung/soft-tissue components. Small isolated groups can project as spurious focal opacities in the bone target. Refinement (Fig.~\ref{fig:abl_boneseg}; Table~\ref{tab:abl_boneseg}) reduces the median number of connected components per CT from $318$ (threshold) to $201$ (iterative), then $2$ after pruning: a median $99.2\%$ reduction with $99.31\%$ bone volume retained (median deleted volume $8.6$~cm$^3$). The effect is largest in the noisiest thin-slice reconstructions, where counts fall from about $1.4$ million to $5$ and the pruned speckle accounts for $30$--$37\%$ of the pre-pruning bone-labelled volume. Iteration adds $316$~cm$^3$ of bone-labelled volume ($932\rightarrow1248$~cm$^3$), while the fraction of this volume in the largest connected component rises $0.856\rightarrow0.975$; in the median CT, pruning then removes only about $0.7\%$ of the volume. Together with neighbourhood-supported growth (Algorithm~\ref{alg:boneseg}), this suggests recovery of bone margins rather than scattered noise.

\label{sec:abl_inpaint}
Removing bone voxels leaves gaps occupying a mean $2.9\%$ of the soft-tissue volume. We compare six ways to fill these gaps before projection: air, $-50$~HU or water constants; nearest-value propagation; slice-wise Telea inpainting, which propagates nearby image values into the gaps; and our iterative three-dimensional diffusion fill, which repeatedly averages neighbouring density values inside the voids. The bone-edge ratio measures how strongly residual edges in the soft-tissue projection coincide with bone edges; lower values indicate fewer bone-shaped imprints. The rim ratio applies the same test to the strongest bone-edge regions. Diffusion has the lowest bone-edge ratio among these fills (Table~\ref{tab:abl_inpaint}); constant $-50$~HU is marginally better on rim ratio ($0.990$ versus $0.992$). Air/water raise bone-edge ratio by $0.159/0.051$ and yield rim ratios $1.320/1.046$, reflecting rib-shaped imprints. Diffusion improves the mean bone-edge ratio over Telea, nearest-value and $-50$~HU fills by $0.012$--$0.016$ across $5{,}000$ CTs; per-CT SDs are about $0.07$.

\label{sec:abl_bs}
Finally, we compare suppression-model architectures trained on identical targets: our single-pass model uses a Swin Transformer V2 (SwinV2) encoder~\cite{swinv2} to extract image features and a UNet++ decoder~\cite{unetpp} to reconstruct the component image. The alternative is a diffusion-transformer backbone (DiT-B/2)~\cite{dit} trained by flow matching~\cite{flowmatching}, which learns iterative updates from noise to the target component. The single-pass suppressor outperforms the flow-matching variant on TBX11K by $0.042$ CPM for RetinaNet and $0.016$ for Faster R-CNN, with a larger Node21 decline for flow (Table~\ref{tab:arch}). The flow-matching model was trained at $512^2$ because $1024^2$ was computationally prohibitive. On an NVIDIA L40S GPU, single-pass suppression takes $0.099$~s per $1024^2$ image ($10.1$ images/s); flow achieves $1.6$ to $0.1$ images/s at $50$--$1000$ steps, $6$--$110\times$ slower with twice the parameters and half the resolution (Table~\ref{tab:runtime}). This cost also matters when constructing translation targets at scale.

\FloatBarrier
\section{Discussion}
\label{sec:discussion}
Anatomically decomposed CT projections provide spatially aligned supervision for chest-radiograph suppression without paired real acquisitions. Bone-mask refinement, soft-tissue inpainting and component-wise rendering generate additive training pairs. Models trained on these synthetic pairs transfer to real radiographs without dataset-specific adaptation, supporting bone suppression and sequential anatomical decomposition. Component-wise translation extends the framework to projections with improved radiograph-like appearance.

Across TBX11K, Node21 and VinDr-CXR, our bone-suppressed-only inputs achieve the highest mean CPM and mAP@50 among the evaluated suppression methods with both detectors. The largest gains over original-image inputs occur on TBX11K. Node21 benefits more from combining the original and suppressed images, and on VinDr our combined input improves mean CPM and mAP@50 over the original-image baseline with both detectors while ranking first among suppression methods. Combined input improves mean CPM over original images in all six settings. These findings support both direct use of our suppressed images and their use as a complementary detector input; the preferred configuration depends on the task and detector.

Lesion conspicuity provides complementary evidence: our method increases nodule contrast on Node21 and VinDr, with confidence intervals above zero, while the evaluated DES-trained diffusion methods reduce contrast in these analyses. Together with the clear-lung detail measurements, these findings support the utility of predicting and subtracting bone. CXR-BS achieves a comparable removal--retention balance on Node21 and VinDr at its native resolution, but has weaker lesion-conspicuity and detection results. The combined evaluations distinguish suppression quality beyond image appearance alone.

The overlap analyses examine whether suppression benefits concentrate in lesions obscured by bone. Recall gains favour overlapped Node21 nodules, while VinDr combined input improves both overlapped and clearer findings. Our suppressed-only input also gives the highest mean recall in both groups in five of six dataset--detector settings. Thus, the comparative detection advantage extends beyond one overlap subgroup. Positive class-level associations on VinDr motivate further investigation of bone overlap as a factor in benefit, alongside lesion characteristics and detector behaviour.

Component-wise translation improves agreement with real-radiograph distributions while leaving the projected lung component unchanged. Lower feature, spectral, entropy and cardiothoracic-ratio distances accompany increased anatomical-structure detection and improved source-CT lung-field agreement. Transfer-probe AUROC remains similar to that of the base projections, although DeepDRR achieves the highest value. These evaluations demonstrate improved image appearance and measured anatomical agreement, with finding discrimination assessed separately.

The ablations support the pipeline design. Bone-mask refinement removes isolated regions while retaining most bone-labelled volume, and diffusion inpainting reduces bone-shaped residual edges. With the same targets, the single-pass suppressor achieves higher detection scores and greater throughput than the evaluated flow-matching model. The comparison assesses two implemented configurations whose different training resolutions prevent attributing the full difference to architecture. Comparisons with released suppression systems likewise assess complete pipelines with different training data and implementation choices.

\label{sec:limitations}
Several constraints define the scope of these findings. Small overlap subgroups limit precision, and individual Node21 subgroup differences remain within image-level sampling uncertainty. Detector results use three training seeds; VinDr Faster R-CNN shows greater between-seed variability, and class-level correlations are exploratory and uncorrected for multiple testing. Class frequency, lesion characteristics and dataset differences limit causal interpretation of the overlap patterns; correlations were not directly compared between methods.

Image-quality measures assess contrast, texture and global anatomical agreement rather than reader performance or complete lesion-level fidelity. Projection detail is constrained by source-CT resolution and the $512^2$ rendering grid. The lung-component extractor is assessed through decomposition and translation, without an independent downstream evaluation, and component errors can propagate into the outputs. Lesion-level fidelity studies, prospective reader evaluations and diagnostic-model training with translated projections would extend the present evidence.

The framework is intended for researchers developing chest-radiograph analysis methods, with predicted components serving as auxiliary representations of the original image. Anatomical CT projection provides useful suppression supervision and supports component-wise translation, connecting CT anatomy with the need for aligned training targets and controllable synthetic radiographs.

\section{Methods}
\label{sec:methods}

This retrospective study develops and evaluates anatomical projection and image-suppression methods using existing public CT and chest-radiograph datasets.

\subsection{Stage 1: anatomical decomposition and projection}
\label{sec:stage1}
\label{par:boneseg}\label{par:decomp}
Stage 1 comprises anatomical decomposition, per-tissue rendering, and additive compositing. Algorithm~\ref{alg:pipeline} summarises the complete pipeline.

\subsubsection{Bone segmentation and anatomical decomposition}
Algorithm~\ref{alg:boneseg} specifies bone segmentation and component construction. Iterative threshold relaxation recovers lower-density bone margins, while neighbourhood support and connected-component pruning reject isolated dense non-osseous structures. The first pass is unfiltered ($M_0=\mathbf1$); subsequent passes require support from the preceding mask, with high-specificity reference voxels retained unconditionally. Pruned voxels return to the soft-tissue component; the size-based pruning can also remove small ossifications.

A pretrained lung segmenter~\cite{lungseg} separates lung (vessels, parenchyma and intrapulmonary findings) from non-lung soft tissue. The masks partition the CT into anatomical regions. Density-domain diffusion then estimates soft-tissue density within the removed-bone regions to prevent rib-shaped void imprints. The resulting component volumes therefore include both the bone and its estimated soft-tissue replacement.

\begin{algorithm}[!htb]
\caption{Anatomy-aware bone segmentation and three-component splitting}
\label{alg:boneseg}
\begin{algorithmic}[1]
\Require CT volume $H$; threshold sequence $\tau=(200,150,100)$~HU;
  reference threshold $\tau_{\mathrm{ref}}=300$~HU; neighbourhood window
  $\mathcal N$ ($15^3$ voxels) with all-ones kernel $\mathbf 1_{\mathcal N}$;
  support ratio $\theta=0.05$; island fraction $\kappa=0.01$; soft-tissue fill
  $v_{\mathrm{fill}}=-50$~HU; structuring elements $S_1{=}\mathrm{ball}(1)$, $S_2{=}\mathrm{ball}(2)$
\Ensure component volumes $H_{\mathrm{bone}}$, $\rho_{\mathrm{NL}}$ (inpainted non-lung
  soft-tissue density), $H_{\mathrm{LN}}$ (lung)
\State $M_{\mathrm{ref}} \gets [\,H > \tau_{\mathrm{ref}}\,]$
  \Comment{high-specificity reference bone}
\State $M_0 \gets \mathbf{1}$
  \Comment{all voxels: first pass is unfiltered}
\Statex \hspace{-\algorithmicindent}\emph{-- iterative threshold relaxation --}
\For{$i = 1 \ldots |\tau|$}
  \State $R \gets \textsc{Close}\big(\lnot[\,H > \tau_i\,],\, S_1\big)$
    \Comment{close background: absorb thin diaphragm}
  \State $M^{\mathrm{raw}}_i \gets \textsc{Close}\big(\lnot R,\, S_2\big)\ \lor\ M_{\mathrm{ref}}$
    \Comment{fill bone interior; restore reference}
  \State\label{eq:support} $s_i \gets \big(M_{i-1} * \mathbf 1_{\mathcal N}\big)\,/\,|\mathcal N|$
    \Comment{fraction of accepted neighbours}
  \State\label{eq:update} $M_i \gets \big(M^{\mathrm{raw}}_i \wedge [\,s_i \ge \theta\,]\big)\ \lor\ M_{\mathrm{ref}}$
    \Comment{retain supported or reference voxels}
  \State store $M_i$ as $\mathcal M[\tau_i]$
\EndFor
\Statex \hspace{-\algorithmicindent}\emph{-- read-off and component pruning --}
\State $M_{\mathrm{bone}} \gets \mathcal M[100]$
  \Comment{bone mask at the final threshold}
\State $\{c_j\} \gets \textsc{ConnectedComponents}(M_{\mathrm{bone}})$;\quad
       $n_{\max} \gets \max_j |c_j|$
\State $M_{\mathrm{bone}} \gets M_{\mathrm{bone}} \setminus
       \{\,c_j : |c_j| < \kappa\,n_{\max}\,\}$
  \Comment{prune small isolated components}
\State $M_{\mathrm{soft}} \gets \lnot M_{\mathrm{bone}}$
  \Comment{soft tissue = complement of the pruned bone mask}
\Statex \hspace{-\algorithmicindent}\emph{-- component volumes --}
\State $H_{\mathrm{bone}} \gets H \odot M_{\mathrm{bone}}
       + H_{\min}\odot(\lnot M_{\mathrm{bone}})$
  \Comment{bone: CT within mask, $H_{\min}{=}\min(H)$ elsewhere}
\State $H_{\mathrm{soft}} \gets H \odot M_{\mathrm{soft}}
       + v_{\mathrm{fill}}\,(\lnot M_{\mathrm{soft}})$
  \Comment{remove bone ($v_{\mathrm{fill}}{=}-50$; refilled below)}
\Statex \hspace{-\algorithmicindent}\emph{-- lung split out of the soft tissue --}
\State $M_{\mathrm{LN}} \gets \textsc{LungSegment}(H)$
  \Comment{pretrained lung model~\cite{lungseg}}
\State $H_{\mathrm{LN}} \gets H_{\mathrm{soft}} \odot M_{\mathrm{LN}}
       + H_{\min}\odot(\lnot M_{\mathrm{LN}})$
  \Comment{lung/vascular component}
\State $H_{\mathrm{NL}} \gets H_{\mathrm{soft}} \odot (\lnot M_{\mathrm{LN}})
       + H_{\min}\odot M_{\mathrm{LN}}$
  \Comment{non-lung soft tissue}
\Statex \hspace{-\algorithmicindent}\emph{-- bone-void inpainting (density domain) --}
\State $\rho_{\mathrm{NL}} \gets \textsc{HuToDensity}(H_{\mathrm{NL}})$
  \Comment{density mapping in Eq.~\eqref{eq:transfer}}
\State $\rho_{\mathrm{NL}} \gets \textsc{DiffusionInpaint}\big(\rho_{\mathrm{NL}},\, M_{\mathrm{bone}}\big)$
  \Comment{voids seeded at $0$; iterative $3^3$ mean diffusion to convergence}
\State \Return $H_{\mathrm{bone}},\, \rho_{\mathrm{NL}},\, H_{\mathrm{LN}}$
  \Comment{$\rho_{\mathrm{NL}}$ enters Alg.~\ref{alg:pipeline} directly as a density field}
\end{algorithmic}
\end{algorithm}

\subsubsection{Density conversion and bone-void inpainting}
After removing the CT bed and clipping HU values to $[-1000,2000]$, we map each voxel value $H(v)$ to a density surrogate using
\begin{equation}
\rho_c(v)=f(H(v)),\qquad f(h)=\frac{3}{1+\exp(-(h-50)/200)}\ \mathrm{g/cm^3}.
\label{eq:transfer}
\end{equation}
This smooth, increasing sigmoid assigns low density to air and approaches $3$~g/cm$^3$ for dense bone. Its midpoint is $50$~HU and transition width $200$~HU. The same bounded mapping is used for all components; tissue-specific attenuation enters through $\mu_c(E)$. We use this mapping as an empirical density surrogate without phantom calibration. Bone-void inpainting acts on $\rho_{\mathrm{NL}}$ after conversion, so it fills the quantity subsequently integrated by the projector.

\label{app:inpaint}
Algorithm~\ref{alg:inpaint} fills bone voids in $\rho_{\mathrm{NL}}$ while retaining known voxels. Diffusing density directly avoids passing averaged HU through the nonlinear transfer function, which can leave intensity steps at bone boundaries.

\begin{algorithm}[!htb]
\caption{Iterative diffusion inpainting of bone voids (density domain)}
\label{alg:inpaint}
\begin{algorithmic}[1]
\Require density volume $\rho$ (bone voxels already removed); void mask $V$ (the pruned bone mask); tolerance $\epsilon{=}10^{-4}$; maximum iterations $N_{\mathrm{it}}{=}2000$; check interval $k{=}20$
\Ensure inpainted density volume $\rho$
\State $\rho \gets \rho\odot(\lnot V)$ \Comment{seed the voids at $0$}
\State $K \gets \tfrac{1}{27}\mathbf 1_{3\times3\times3}$ \Comment{box (mean) kernel}
\For{$i = 1 \ldots N_{\mathrm{it}}$}
  \State $\rho_{\mathrm{prev}} \gets \rho$
  \State $\rho \gets (\rho * K)\odot V + \rho_{\mathrm{prev}}\odot(\lnot V)$ \Comment{diffuse inside the voids; restore known voxels}
  \If{$i \bmod k = 0$ \textbf{and} $\mathrm{mean}_{v\in V}\,|\rho(v)-\rho_{\mathrm{prev}}(v)| < \epsilon$}
    \State \textbf{break} \Comment{converged}
  \EndIf
\EndFor
\State \Return $\rho$
\end{algorithmic}
\end{algorithm}

\subsubsection{Per-tissue polychromatic rendering}
\label{par:hu2lac}\label{par:poly}\label{ssec:proj_params}\label{sssec:transfer}
We render bone, lung and non-lung soft tissue separately to obtain rendered images $R_{\mathrm{BN}}$, $R_{\mathrm{LN}}$ and $R_{\mathrm{NL}}$, before the composition weights are applied. Each component $c$ has a spatial density field $\rho_c$ and a tissue-specific mass attenuation coefficient $\mu_c(E)$, which describes attenuation per unit mass at photon energy $E$. Their product gives the linear attenuation coefficient. Rendering separates geometry---how much tissue lies along each ray---from the energy-dependent attenuation of that tissue.

A Siddon--Jacobs ray integrator~\cite{siddon1985,jacobs1998} sums density times path length through the voxels intersected by each ray, accounting for voxel spacing. Each component is projected using the same source-detector geometry and a $512\times512$ detector grid, ensuring spatial alignment of the component projections. The resulting area density at detector pixel $(x,y)$ is
\begin{equation}
t_c(x,y)=\int_{\mathrm{ray}(x,y)}\rho_c\,\mathrm ds\quad(\mathrm{g/cm^2}),
\label{eq:areadensity}
\end{equation}
the mass of component $c$ per unit area along that ray. This geometric projection does not yet include attenuation.

\emph{Tissue attenuation, source spectrum and detector response.} Each component receives an elemental composition from the International Commission on Radiation Units and Measurements (ICRU) Report~44 reference tissues~\cite{icru44} (Table~\ref{tab:materials}). Algorithm~\ref{alg:mu} combines elemental cross-sections from the National Institute of Standards and Technology (NIST) XCOM database~\cite{xcom} by mass fraction to obtain $\mu_c(E)$ in cm$^2$/g. This accounts for composition differences such as calcium and phosphorus in bone. SpekPy~\cite{spekpy} provides the incident photon fluence $\Phi(E)$ at discrete energies for the selected tube potential, using a tungsten anode, $12^\circ$ anode angle and $2.5$~mm aluminium filtration.

\label{app:mu}
Algorithm~\ref{alg:mu} applies the mass-fraction mixture rule to the compositions in Table~\ref{tab:materials}, converting elemental cross-sections from cm$^2$/atom to cm$^2$/g.

\begin{algorithm}[!htb]
\caption{\textsc{MaterialMu}: attenuation from mass fractions}
\label{alg:mu}
\begin{algorithmic}[1]
\Require atomic numbers $Z=(z_1,\ldots,z_n)$; mass fractions $f=(f_1,\ldots,f_n)$;
  energies $E$; Avogadro constant $N_A$; standard atomic mass $A(z)$ (g/mol)
\State $\mu(E) \gets 0$
\For{$j = 1 \ldots n$}
    \State $\sigma_j(E) \gets \textsc{XcomCrossSection}(z_j, E)$ \Comment{total photon cross-section per atom}
    \State $(\mu/\rho)_j(E) \gets \sigma_j(E)\cdot N_A / A(z_j)$ \Comment{per-atom cross-section $\rightarrow$ per-gram, cm$^2$/g}
    \State $\mu(E) \gets \mu(E) + f_j\,(\mu/\rho)_j(E)$ \Comment{mixture rule}
\EndFor
\State \Return $\mu(E)$
\end{algorithmic}
\end{algorithm}

At energy $E$, the Beer--Lambert law gives the transmitted fraction $\exp[-\mu_c(E)s_c(V)t_c(x,y)]$, where $s_c(V)$ is the dimensionless component scale defined in Eq.~\eqref{eq:component_scale} and $V$ is the numerical tube potential in kVp. An energy-integrating detector weights surviving photons by their energy, giving the factor $E$ below. We sum over energies, divide by the unattenuated detector signal (the flat field), and take the negative logarithm:
\begin{equation}
L_c(x,y)=-\log\!\frac{\sum_E\Phi(E)E\exp[-\mu_c(E)s_c(V)t_c(x,y)]}{\sum_E\Phi(E)E}.
\label{eq:render}
\end{equation}
The denominator is the signal with no attenuating tissue. Thus $L_c=0$ on an unattenuated ray and increases with tissue mass. Preferential absorption of lower-energy photons produces beam hardening, with different effects for each tissue composition. Changing tube potential changes $\Phi(E)$ and therefore component contrast without requiring another geometric projection.

Equation~\eqref{eq:render} describes the noise-free signal. In practice, low-frequency scatter, Poisson photon noise and detector blur are applied to transmission before the logarithm. The resulting log image is min--max scaled to give the rendered component $R_c$. The component scales below incorporate any tube-potential multiplier; composition weights are applied subsequently to $R_c$.

\subsubsection{Calibration and projection sweep}
\label{sssec:calibration}
The empirical component scales applied inside the exponential in Eq.~\eqref{eq:render} are
\begin{equation}
s_c(V)=
\begin{cases}
11.667, & c=\mathrm{LN},\\
10.8\,(V/60), & c=\mathrm{NL},\\
3.0\,(V/60), & c=\mathrm{BN}.
\end{cases}
\label{eq:component_scale}
\end{equation}
The lung scale is constant across tube potentials and receives no additional $V/60$ multiplier. Lung is additionally exported at $3\times$ scale for translated recombination. Empirical scales are fixed once. The noise model comprises low-frequency scatter (Gaussian $\sigma=20$~px, scatter-to-primary ratio $0.04$), Poisson noise ($N_0=10^6$ photons/pixel, counts floored at one) and detector blur ($\sigma=0.8$~px). The suppression corpus varies tube potential and PA/AP view; the CT-RATE quality corpus uses PA views at $\{80,100,120,150\}$~kVp, with all quality metrics computed at $100$~kVp.

\subsubsection{Additive compositing}
\label{sec:compositing}
We distinguish the rendered component image $R_c$ from its weighted contribution $P_c$ to the base DRR:
\begin{equation}
P_c=w_cR_c,\qquad c\in\{\mathrm{BN},\mathrm{NL},\mathrm{LN}\}.
\label{eq:weighted_component}
\end{equation}
The fixed composition weights are $w_{\mathrm{BN}}=1$, $w_{\mathrm{LN}}=1.75/1.3\approx1.346$ and $w_{\mathrm{NL}}=3.5/1.3\approx2.692$, chosen once to obtain radiograph-like contrast. These weights act on rendered images, separately from the density scales applied before attenuation (Sec.~\ref{sssec:calibration}). The base DRR, \base{}, and its bone-suppressed counterpart are
\begin{equation}
P_{\mathrm{full}}=P_{\mathrm{BN}}+P_{\mathrm{NL}}+P_{\mathrm{LN}},\qquad
P_{\mathrm{ST}}=P_{\mathrm{NL}}+P_{\mathrm{LN}}.
\label{eq:additive}
\end{equation}
Thus $P_c$ consistently denotes a contribution with its composition weight already included. During suppression training, the lung and non-lung soft-tissue weights are sampled per image, and the same normalisation factor is applied to the assembled image and its component contributions (Sec.~\ref{ssec:bs_config}). Registered, additive targets support prediction of one component and subtraction to recover its complement. This is a physics-informed supervision model: components are independently noise-corrupted and log-transformed, and both bone and its inpainted replacement contribute. It therefore differs from intact-chest polychromatic transmission, where attenuation combines within a single exponential.

\paragraph{Pipeline summary}
\label{app:pipeline}
Algorithm~\ref{alg:pipeline} assembles the decomposition, inpainting, attenuation and compositing steps described above.

\begin{algorithm}[!t]
\caption{Anatomy-aware base projection (Stage~1)}
\label{alg:pipeline}
\begin{algorithmic}[1]
\Require CT volume $H$; tube potentials $\mathcal K$; views $\mathcal V$; photon count $N_0$; scatter-to-primary ratio $\eta$
\State $H \gets \textsc{RemoveBed}(H)$;\quad clip $H$ to $[-1000,2000]$~HU
\State $(H_{\mathrm{bone}}, \rho_{\mathrm{NL}}, H_{\mathrm{LN}}) \gets \textsc{AnatomySplit}(H)$ \Comment{Alg.~\ref{alg:boneseg}: bone seg, inpainting, lung split}
\State $\rho_{\mathrm{BN}} \gets \textsc{HuToDensity}(H_{\mathrm{bone}})$;\quad $\rho_{\mathrm{LN}} \gets \textsc{HuToDensity}(H_{\mathrm{LN}})$ \Comment{Eq.~\eqref{eq:transfer}; $\rho_{\mathrm{NL}}$ is already an (inpainted) density field}
\ForAll{$(kVp, \text{view}) \in \mathcal K \times \mathcal V$}
    \State $(E,\Phi) \gets \textsc{SpekPySpectrum}(kVp)$ \Comment{energies, relative fluence}
    \ForAll{$c$}
        \State $t_c \gets \textsc{Project}(\rho_c, \text{view})$ \Comment{Siddon--Jacobs line integral}
        \State $s_c \gets s_c(kVp)$ \Comment{Eq.~\eqref{eq:component_scale}; lung has no extra $kVp/60$ factor}
        \State $\mu_c(E) \gets \textsc{MaterialMu}(Z_c, f_c, E)$ \Comment{Alg.~\ref{alg:mu}}
    \EndFor
    \ForAll{$c$}
        \State $T_c \gets \textstyle\sum_E \Phi(E)E\,e^{-\mu_c(E)s_ct_c}\big/\sum_E \Phi(E)E$ \Comment{per-component transmission}
        \State apply \textsc{Scatter}, \textsc{PoissonNoise}$(\cdot,N_0)$, \textsc{DetectorBlur} to $T_c$ \Comment{before the log (Sec.~\ref{par:poly})}
        \State $L_c \gets -\log T_c$;\quad $R_c \gets \textsc{MinMaxScale}(L_c)$ \Comment{rendered component before composition weighting}
    \EndFor
    \State $P_c \gets w_cR_c$ for each component $c$ \Comment{composition weights, Eq.~\eqref{eq:weighted_component}}
    \State $P_{\mathrm{full}} \gets P_{\mathrm{BN}}+P_{\mathrm{NL}}+P_{\mathrm{LN}}$ \Comment{Eq.~\eqref{eq:additive}}
    \State save rendered components $R_c$ and the assembled base image $P_{\mathrm{full}}$
\EndFor
\end{algorithmic}
\end{algorithm}

\subsection{Stage 2: component prediction and subtraction}
\label{sec:uses}\label{sec:lungcomp}\label{ssec:bs_config}
The bone suppressor predicts $\hat P_{\mathrm{BN}}$ from $P_{\mathrm{full}}$; subtraction gives $\hat P_{\mathrm{ST}}=P_{\mathrm{full}}-\hat P_{\mathrm{BN}}$. Bone's regular appearance motivates predicting this component, and subtraction confines the image modification to the predicted residual. A second model predicts the entire lung component from soft tissue, giving $\hat P_{\mathrm{NL}}=P_{\mathrm{ST}}-\hat P_{\mathrm{LN}}$. Operating after bone removal simplifies this extraction. This model supplies the real-radiograph components used as translation targets.

Both models use an ImageNet-pretrained SwinV2-Base encoder~\cite{swinv2} (window~8, four stages), UNet++ decoder~\cite{unetpp} (channels $256/128/64/32$), one $1024^2$ input channel, and a prediction head comprising bilinear upsampling, a convolutional block attention module (CBAM)~\cite{cbam} and a $1\times1$ convolution. Optimisation settings are in Table~\ref{tab:supp_bs_config}.

\paragraph{Loss}
The objective is
\begin{equation}
\mathcal L \;=\; 0.5\,\mathrm{L1} \;+\; 0.5\,\mathrm{MSE} \;+\;
0.5\,\bigl(1-\mathrm{SSIM}\bigr) \;+\;
w_M\bigl(\mathrm{L1}_M+\mathrm{MSE}_M\bigr),
\label{eq:supp_bs_loss}
\end{equation}
where L1 and MSE denote mean absolute error and mean squared error, respectively. Their equally weighted sum defines the pixel-distance term; the structural similarity index (SSIM) uses an
$11{\times}11$ Gaussian window ($\sigma{=}1.5$, unit data range), and
$\mathrm{L1}_M/\mathrm{MSE}_M$ are restricted to the lung-structure mask $M_{\mathrm{roi}}$ with weight
$w_M{=}2.5$. $M_{\mathrm{roi}}$ is obtained by contrast-sharpening the lung projection
$P_{\mathrm{LN}}$ (contrast-limited adaptive histogram equalisation, CLAHE, followed by unsharp masking) and thresholding at $0.3$ of
its maximum, emphasising the retention of vessels and intrapulmonary findings during bone suppression. The lung-component suppressor uses the same objective without the marking
term ($w_M{=}0$).

\paragraph{On-the-fly pair assembly and augmentation}
Each training pair is assembled from the component projections at load time
(Section~\ref{sec:compositing}). (i)~\emph{Component mixing:} the full image is composited
as $w_{\mathrm{LN}}R_{\mathrm{LN}}+w_{\mathrm{NL}}R_{\mathrm{NL}}+R_{\mathrm{BN}}$ with per-image weights
$w_{\mathrm{LN}}\sim\mathcal U[1.154,\,1.538]$ and $w_{\mathrm{NL}}\sim\mathcal U[1.538,\,3.846]$, then
max-normalized; the target is recomputed from the same weights so the additive
identity holds exactly. At evaluation the fixed weights $w_{\mathrm{LN}}{=}1.75/1.3$,
$w_{\mathrm{NL}}{=}3.5/1.3$ are used (Sec.~\ref{sec:compositing}). For lung-component suppression the
composite is $w_{\mathrm{LN}}R_{\mathrm{LN}}+w_{\mathrm{NL}}R_{\mathrm{NL}}$, and with probability $0.25$
a bone image scaled by $u^4$, $u\sim\mathcal U[0,1]$, is added back.
(ii)~\emph{Appearance:} with probability $0.5$ the full/soft-tissue pair is
histogram-matched (quantile lookup) to a randomly drawn CheXpert CXR~\cite{chexpert}; otherwise a
serial contrast stack is applied: inverse-gamma with
$\gamma\sim\mathcal U[2,3]$, gamma with $\gamma\sim\mathcal U[1.5,3.5]$, then
(with probability $0.6/0.4$) a gamma $\mathcal U[0.5,1.5]$ or inverse-gamma
$\mathcal U[1.5,2.5]$. The lung-component suppressor omits histogram matching and applies
the contrast stack to every sample. After every appearance transform the target
is recomputed by subtraction. (iii)~\emph{Geometry:} rotation up to $90^\circ$
with border cropping ($p{=}0.4$) and random resized crops (area scale
$[0.49,1.0]$, aspect $[0.75,1.33]$, $p{=}0.4$), replayed identically on all
images of the pair, followed by a resize to $1024^2$. Training draws $64{,}000$
samples per epoch with replacement from the variant pool.

\paragraph{Histogram matching and additivity}
For histogram-matched samples, one monotone quantile lookup $T$ maps the full image to a random real frontal CXR from CheXpert~\cite{chexpert} using 256-bin, 8-bit histograms. Apply this same lookup to full and soft-tissue images, then recompute bone as $T(P_{\mathrm{full}})-T(P_{\mathrm{ST}})$. This retains exact additivity and a non-negative target, zero wherever the original bone projection is zero. The nonlinear target is approximately $T'(P_{\mathrm{ST}})P_{\mathrm{BN}}$, rather than $T(P_{\mathrm{BN}})$: its amplitude follows local display contrast, retaining ribs over the steeper lung-field tones and attenuating spine over the saturating mediastinum. Non-matched pairs likewise share each contrast exponent before target subtraction. Real CXRs enter suppression training only through histogram matching (Supplementary Fig.~\ref{supp-fig:supp_aug}).

\subsection{Stage 3: unpaired component-wise translation}
\label{sec:translation}\label{ssec:translator_config}
Separate generators translate synthetic bone and non-lung soft tissue to their model-derived real counterparts, produced by sequential Stage-2 suppression. Restricting translation only to bone and non-lung soft-tissue components narrows each appearance distribution and leaves the fragile synthetic lung markings unchanged.

We adapt SANTA, an unpaired image-translation framework with a shared-decoder shortest-path formulation~\cite{santa}: $G(z,d)$ traces a path from source ($d=0$) to target ($d=1$), with a decoder-feature Jacobian penalty encouraging a short, content-preserving mapping and discouraging mode collapse. We replace SANTA's residual-network generator with adaptive instance normalisation (AdaIN) and plain least-squares generative adversarial network (LSGAN) discriminator with a frozen pretrained MedVAE autoencoder~\cite{medvae}, trainable low-rank adaptation (LoRA) modules~\cite{lora}, latent style/noise conditioning and a vision-aided discriminator~\cite{visionaided} using features from MedCLIP, a medical image--text model held fixed during translation training~\cite{medclip}. The two translators share the following configuration and train independently.

\paragraph{Generator}
The generator uses a pretrained two-dimensional MedVAE
autoencoder~\cite{medvae} with single-channel input and output,
fourfold spatial downsampling and four latent channels.
A $512\times512$ input therefore yields a
$128\times128\times4$ latent representation. The backbone is frozen, with trainable LoRA adapters (rank~$8$, $\alpha{=}8$, dropout $0.05$) injected into the deeper encoder stages and the full decoder.
The domain scalar $d$ conditions the latent through a feature-wise linear modulation (FiLM) layer,
$z_d = (1+s(d))\,z + t(d)$, with $(s,t)$ produced by a three-layer multilayer perceptron (MLP) of width
$8$; during training the latent is perturbed with Gaussian noise
($\sigma{=}0.25$).

\paragraph{Discriminator}
A frozen MedCLIP Swin-Tiny backbone (inputs resized to $224^2$) feeds a trainable
multi-level, spectrally normalised head over three feature levels. The generative adversarial network (GAN)
objective is least-squares with one-sided label smoothing (real target $0.9$),
one discriminator step per generator step, and a $50$-image history pool.

\paragraph{Objective}
\begin{equation}
\mathcal L_G=\mathcal L_{\mathrm{gan}}+5\mathcal L_{\mathrm{rec}}+5\mathcal L_{\mathrm{idt}}+0.01\mathcal L_{\mathrm{kl}}+0.01\mathcal L_{\mathrm{path}}.
\end{equation}
Reconstruction and identity use $\ell_1$ loss; the Kullback--Leibler (KL) surrogate is an $\ell_2$ penalty on the latent mean. The path term penalises squared feature differences at five decoder layers for the same latent decoded at $d_c\pm\delta$, normalised by the scalar gap, where $d_c\sim\mathcal U[0,1]$ and $\delta\sim\mathcal U[0.05,0.10]$. Only this term samples $d$ continuously; reconstruction uses $d=0$, and identity/adversarial losses use $d=1$. The path weight is below SANTA's reported range, and the discriminator learning rate is lower than the generator's.

\paragraph{Schedule and data}
We train the generators and discriminators using Adam with $\beta_1=0.5$ and $\beta_2=0.99$. The generator learning rate is held at $2\times10^{-4}$ for the first four epochs and then decayed linearly to zero over the next six epochs. The discriminator learning rate starts at $1\times10^{-5}$ and decays linearly from the beginning of training. Training uses bfloat16 (bf16) precision and an effective batch size of eight, with one sample per GPU across eight GPUs. We do not apply exponential moving averaging (EMA) to the model weights. The checkpoints are selected by validation generator loss. Training is unpaired at
$512^2$ (Lanczos resize): the source domain is the Stage-1 component projections
and the target domain the corresponding real components produced by the Stage-2
suppressors on real frontal radiographs (Sec.~\ref{sec:uses}). For the
soft-tissue translator, the target is the lung-component-suppressed (non-lung)
soft-tissue component. We hold out $5\%/5\%$ of each domain for validation/test.

\paragraph{Inference and translated recombination}
\label{sssec:recombine}
The released setting decodes the soft-tissue stream at $d{=}0.9$ and the bone stream at $d{=}0.7$. Let $\widetilde R_{\mathrm{NL}}$ and $\widetilde R_{\mathrm{BN}}$ denote the translated component images, and $R_{\mathrm{LN}}^{3\times}$ the untranslated lung image exported with the additional $3\times$ density scale (Sec.~\ref{sssec:calibration}). Recombination uses fixed display weights, distinct from the base-DRR composition weights. First, we form
\begin{equation}
Y=1.6\,\widetilde R_{\mathrm{NL}}+0.6\,R_{\mathrm{LN}}^{3\times}.
\end{equation}
We min--max scale $Y$ to obtain $X\in[0,1]$, subtracting the minimum and dividing by the intensity range. The translated bone image is likewise min--max scaled. We add $0.25$ times this bone image to $X$ and clip the result to $[0,1]$ to obtain \trans{}, the translated radiograph used in the projection-quality comparisons.

\subsection{Datasets and evaluation protocols}
\label{sec:setup}\label{ssec:detcfg}
\label{sec:provenance}\label{sec:ethics}
\paragraph{Ethics and consent}
This retrospective study used existing de-identified public research datasets: CT-RATE~\cite{ctrate}, TBX11K~\cite{tbx11k}, Node21~\cite{node21}, VinDr-CXR~\cite{vindrcxr}, JSRT~\cite{jsrt} and CheXpert~\cite{chexpert}. The datasets were accessed and used in accordance with their respective licences and data-use conditions. No participants were prospectively recruited, no intervention was assigned and no new clinical data were collected. According to the source publications, CT-RATE scans were acquired at Istanbul Medipol University Mega Hospital, Turkey (May 2015 to January 2023), CheXpert radiographs at Stanford Health Care, USA (October 2002 to July 2017), and VinDr-CXR radiographs at Hanoi Medical University Hospital and Hospital 108, Vietnam (2018 to 2020); TBX11K, Node21 and JSRT were used as distributed, and their acquisition dates are not restated here. Source-dataset collection and consent arrangements are described in the corresponding publications and documentation. Additional ethics approval and informed consent were not required for this analysis of previously collected, de-identified public research data.

\subsubsection{Suppression protocol}
CT-RATE contains $25{,}692$ non-contrast CT scans from $21{,}304$ patients with report-derived labels. To generate suppression-training pairs, we selected $6{,}000$ CT volumes prioritising the smallest axial voxel spacing (z-spacing). The selected volumes represented 6,000 patients. For component-predictor training, $5{,}000$ CT volumes from $5{,}000$ patients were used for training and $1{,}000$ volumes from $1{,}000$ patients for validation. Patients in training and validation cohorts are disjoint by construction. The bone and lung-component predictors used the same partitions. For histogram-matching augmentation, each reference radiograph was drawn randomly from CheXpert's frontal chest radiographs; these images supplied intensity-distribution references only and did not participate directly in the supervision.

TBX11K~\cite{tbx11k} has $11{,}200$ radiographs with TB boxes. Our local training and internal-validation partitions contain $5{,}610$ and $990$ images, respectively; evaluation uses the official $1{,}800$-image validation set because official test boxes are withheld. Node21~\cite{node21} has $4{,}882$ frontal images, $1{,}134$ with expert nodule boxes. Following the stratified protocol of Behrendt et al.~\cite{behrendt2023}, our fixed partition contains $3{,}626$ training, $906$ internal-validation and $350$ evaluation images. VinDr-CXR~\cite{vindrcxr} has $18{,}000$ adult frontal images, 17 radiologists and 14 boxed finding classes. Its official $15{,}000$-image training set is divided into $12{,}750$ training and $2{,}250$ internal-validation images; we retain its official $3{,}000$-image consensus test set for evaluation. For TBX11K, both the training and internal-validation partitions were drawn from the official training set, stratified by the presence of TB boxes, with $15\%$ of each stratum reserved for internal validation (random seed $42$); the official validation set was reserved for evaluation. For Node21, nodule-positive and nodule-negative radiographs were shuffled separately (seed $42$), and the evaluation partition contained $175$ nodule-positive and $175$ nodule-negative radiographs drawn from the public dataset. For VinDr-CXR, the subdivision of the official training set was fixed once. Within each dataset, the training, internal-validation and evaluation partitions had no shared image identifiers; patient-level separation could not be verified from the available detector manifests.

For VinDr-CXR, we merge overlapping training-label boxes from the three annotators using weighted boxes fusion~\cite{wbf}. We retain labels supported by at least two radiologists (min\_conf${=}0.4$), giving an annotation density comparable to the consensus test set; both VinDr detectors are initialised with weights pretrained on the Common Objects in Context (COCO) object-detection dataset. JSRT~\cite{jsrt} ($247$ images, $154$ confirmed nodules) is used for lesion-conspicuity, qualitative and preservation analyses; no separate detector is trained on JSRT.

The primary detection evaluation comprises TBX11K, Node21 and VinDr-CXR. We apply the same bone suppressor to all three datasets without adaptation and train the downstream detectors separately within each dataset. RetinaNet and Faster R-CNN use ResNet-50 backbones with feature pyramid networks (R50-FPN-v2) and train on $full$, $bs$ or two-channel $full_{bs}$ inputs, with otherwise identical settings, splits and preprocessing across arms and methods (Table~\ref{tab:detcfg}): $512^2$ for TBX11K, $1024^2$ for Node21/VinDr. Baselines use their strongest released checkpoints at native resolution, without retraining. Seed counts are specified in Sec.~\ref{sec:statistics}.

\subsubsection{Projection protocol}
DeepDRR~\cite{deepdrr}, Plastimatch~\cite{plastimatch}, nanoDRR/DiffDRR (an optimised DiffDRR implementation~\cite{diffdrr}) and TIGRE~\cite{tigre} render the same CTs. Other learned-realism/Gaussian-splatting renderers~\cite{realdrr,xgaussian,ddgsct} lack usable public code or require per-volume optimisation. We exclude softMip~\cite{softmip}: its published weighting, reproduced by Paalvast et al.~\cite{paalvast2025}, omits the normalisation/output mapping needed for reproducible FID comparison. Real references are all $18{,}000$ VinDr images and $191{,}027$ frontal images from $64{,}534$ CheXpert patients~\cite{chexpert}. These pools provide reference image distributions rather than additional independent clinical test cohorts.

Per-image metrics, including structure counts, use a matched set of $22{,}124$ CT-RATE reconstruction volumes with axial spacing $\le1$~mm, corresponding to $17{,}082$ study identifiers and $14{,}538$ patient identifiers. Of $22{,}125$ volumes meeting the spacing criterion, one was excluded: this $1{,}106$-slice acquisition (about $830$~mm long at $0.75$~mm spacing) extends far beyond the thorax, and the body occupies only $12.5\%$ of the volume. This volume was excluded from every comparison arm. CT-referenced geometry uses $21{,}578$ volumes from $14{,}524$ patients whose Stage-1 lung segmentation occupies at least $1\%$ of the padded CT volume. Reconstruction volumes, studies and patients are counted separately. Domain scalars were selected by a validation sweep for realism with retained content; bone tolerated less translation before rib-texture drift.

\subsubsection{Primary metrics}
We evaluate lesion detection using FROC CPM and COCO mAP@50. We use a study-specific FROC protocol in which CPM is the mean sensitivity at $0.125$, $0.25$, $0.5$, $1$, $2$, $4$ and $8$ false positives per image, with detections matched to reference boxes at IoU $\geq 0.3$. At each operating point, we use the highest sensitivity attained without exceeding the specified false-positive rate. This protocol is applied consistently across methods and datasets and differs from the official NODE21 challenge scoring protocol. COCO mAP@50 summarises ranked-box precision across recall using IoU $\geq 0.5$. For the overlap-group comparisons, we select one global threshold per model at $2$ false positives per image and apply it to both lesion groups.

We evaluate distributional realism using FID~\cite{fid} and KID~\cite{kid}, calculated from the $1{,}024$ penultimate features of TorchXRayVision DenseNet-121~\cite{xrv}, pretrained on CXRs (\texttt{all} weights). Images are evaluated at $224^2$, and real and synthetic sample sizes are matched. We average scores over $10$ subsamples; lower values indicate closer agreement between the feature distributions.

\subsection{Lesion overlap and image-preservation metrics}
\label{par:surrogates}
DES-reference FID, KID, peak signal-to-noise ratio (PSNR) and learned perceptual image patch similarity (LPIPS) may reward agreement with the DES training distribution, including motion/misregistration, dose-splitting noise and residual cross-contamination. Published GL-LCM training/evaluation also excludes pneumothorax and pleural effusion~\cite{gllcm}. We therefore prioritise detection and lesion conspicuity over DES-reference similarity as cross-system suppression measures.

We consider lesions as bone-overlapped when at least $15\%$ of their bounding-box pixels intersect the bone map from our suppressor. This common grouping is used across methods. CNR~\cite{rose1948} is $|\mu_{\mathrm{lesion}}-\mu_{\mathrm{bg}}|/\sigma_{\mathrm{bg}}$, using the annotated box and a surrounding annulus excluding it. The CNR analysis includes every lesion whose box lies at least $25\%$ inside the dilated rib and clavicle region of the independent CXAS segmenter~\cite{cxas}; this selection is made once per lesion and is identical for every method, and it is separate from the $15\%$ predicted-bone criterion used for the detection overlap groups. After intensity matching over non-bone lung, we report suppressed-minus-input CNR; positive changes indicate greater local contrast.

For removal/retention, $X$ is the input and $Y$ the suppressed output intensity-matched over clear lung. CXAS~\cite{cxas} also defines bone mask $M_b$ and clear lung $M_c$ (lung minus dilated bone). We measure
\begin{align}
\mathrm{Removal}&=1-\frac{\langle|B(Y)|\rangle_{M_b}}{\langle|B(X)|\rangle_{M_b}},\qquad B=G_{\sigma_{\mathrm{hi}}}*\,\cdot-G_{\sigma_{\mathrm{lo}}}*\,\cdot,\\
\mathrm{Retention}&=\frac{\mathrm{Var}[\nabla^2Y]_{M_c}}{\mathrm{Var}[\nabla^2X]_{M_c}}.
\end{align}
Here $\langle\cdot\rangle_M$ and $\mathrm{Var}[\cdot]_M$ denote masked mean and variance; $B$ is a difference-of-Gaussians rib-edge band-pass~\cite{marrhildreth1980} with $\sigma_{\mathrm{hi}}=w/6$, $\sigma_{\mathrm{lo}}=w/2$, $w=12$~px at $1024^2$, scaled with resolution. Laplacian-variance retention~\cite{pechpacheco2000} targets unity: low values suggest blur; high, variable values suggest added/oversharpened texture.

Upsampling CXR-BS from $256^2$ to $1024^2$ retains only $1$--$2\%$ clear-lung Laplacian variance, a resolution effect. We therefore score ours and diffusion methods at native $1024^2$, and compare CXR-BS with ours at $256^2$, rescaling band-pass and dilation. Supplementary Table~\ref{supp-tab:supp_antiblur256} compares all methods at $256^2$. For each dataset and evaluation resolution, we compare each baseline with our method using two-sided Wilcoxon signed-rank tests, pairing measurements from the same radiographs. Bone-removal and detail-retention scores are tested separately.

\subsection{Projection-fidelity and ablation metrics}
\label{sec:projection_metrics}
We evaluate the synthetic projections at three levels: image appearance, agreement with source anatomy, and the contribution of individual projection steps. Appearance measures compare spatial-frequency content, sharpness and texture with real radiographs, while the anatomical evaluations assess finding discrimination and structural agreement.

For spatial-frequency content, we compute the squared magnitude of each image's two-dimensional Fourier transform and average it within $256$ radial frequency bins at $512\times512$ resolution. We first average these profiles across images within each domain, giving $\bar S_{\mathrm{syn}}(r)$ and $\bar S_{\mathrm{real}}(r)$, and then compare their logarithms:
\begin{equation}
D_{\mathrm{spec}}=\left[\sum_{r=0}^{255}\left\{\log_{10}\!\bigl(\max[\bar S_{\mathrm{syn}}(r),\epsilon_s]\bigr)-\log_{10}\!\bigl(\max[\bar S_{\mathrm{real}}(r),\epsilon_s]\bigr)\right\}^{2}\right]^{1/2},
\label{eq:spectral_distance}
\end{equation}
where $\epsilon_s=10^{-30}$ prevents taking the logarithm of zero. Lower distance indicates closer agreement in spatial-frequency content. We complement this distribution-level measure with per-image Laplacian variance~\cite{pechpacheco2000}, computed over the whole image and within the lung mask predicted by the independent anatomy segmenter described below.

To assess texture, we compute local entropy from the greyscale intensity histogram within a disk of radius $5$ pixels and average the resulting entropy map within a heuristic lung-field region. This region comprises pixels below $40\%$ of the maximum intensity in the central $80\%$ of the image; images with fewer than $100$ eligible pixels are omitted from this metric. We then calculate the Wasserstein distance between the synthetic and real distributions of per-image mean entropy. This local-entropy measure is distinct from entropy derived from a grey-level co-occurrence matrix. Both the spectral and texture analyses use $512\times512$ images, with lower distances indicating closer agreement with real-image distributions.

We assess whether the projections retain information about findings in the source CT using a frozen classifier trained only on real CheXpert CXRs. Patient identifiers were shuffled (seed $42$) and assigned approximately $70\%/15\%/15\%$ to training, validation and test, keeping each patient's radiographs together. The resulting patient-disjoint partitions contained $133{,}035$, $29{,}120$ and $28{,}872$ frontal radiographs from $45{,}174$, $9{,}680$ and $9{,}680$ patients, respectively. Checkpoint selection used a fixed subset of $6{,}000$ validation radiographs from $3{,}721$ patients. The classifier was then held fixed when scoring the generated projections. The full frontal-image pool also supplies the CheXpert realism reference. Its predictions are evaluated against source-CT labels for pleural effusion, atelectasis, cardiomegaly and consolidation, and AUROC-4 is the arithmetic mean of their four AUROCs. These findings have corresponding label definitions across the two modalities. We exclude lung opacity and lung lesion because the CT definitions include ground-glass/density increase and fissural nodules, respectively, without equivalent radiographic definitions.

We complement this finding-level evaluation with an independent anatomical assessment using the TorchXRayVision~\cite{xrv} PSPNet~\cite{pspnet} trained on ChestX-Det~\cite{chestxdet}. A structure is counted as detected when its predicted probability exceeds $0.5$ at any pixel, giving a count out of $14$ structures per image. Thresholding the heart and lung probability maps at $0.5$ also supplies the masks used to calculate cardiothoracic ratio: the horizontal extent of the heart divided by that of the union of both lungs. We compare synthetic and real cardiothoracic-ratio distributions using Wasserstein distance. Finally, to assess agreement with the individual source CT, we calculate IoU between the predicted union of the two lung masks and the rectangular region obtained by projecting that CT's lung bounding box onto the image plane. The CT boxes are derived from our Stage-1 lung segmentation on the same padded HU volume used for projection; this comparison includes only volumes with at least $1\%$ segmented lung voxels.

The projection ablations examine how bone-mask refinement and void filling affect the synthetic targets. All variants use the same fixed subset of $5{,}000$ CT-RATE volumes from $5{,}000$ patients with z-spacing $0.50$--$1.00$~mm, projected in the PA view at $100$~kVp; all $5{,}000$ of these volumes also occur in the projection-evaluation set. For bone segmentation, we successively add iterative threshold relaxation and neighbourhood filtering to a fixed HU threshold, followed by connected-component pruning. Changes in mask volume and connected-component counts quantify how these steps retain bone-labelled tissue while removing isolated regions; voxel-level bone annotations are unavailable in CT-RATE.

For void filling, we compare air ($-1000$~HU), $-50$~HU and water ($0$~HU) constants with nearest-value propagation by Euclidean distance transform, slice-wise Telea inpainting and three-dimensional diffusion. We evaluate the residual bone-shaped edges in each resulting soft-tissue projection using two complementary ratios. Let $g_{\mathrm{ST}}$ and $g_{\mathrm{BN}}$ denote the gradient magnitudes of the soft-tissue and bone projections, computed using $3\times3$ Sobel filters on intensities scaled to $[0,1]$, and let $\Omega$ denote the body region. The bone-edge ratio is
\begin{equation}
R_{\mathrm{edge}}=
\frac{\displaystyle\sum_{p\in\Omega}g_{\mathrm{ST}}(p)g_{\mathrm{BN}}(p)}
{\displaystyle\sum_{p\in\Omega}g_{\mathrm{BN}}(p)+\epsilon_g}
\frac{1}{\langle g_{\mathrm{ST}}\rangle_{\Omega}+\epsilon_g},
\label{eq:bone_edge_ratio}
\end{equation}
where $\epsilon_g=10^{-9}$ and $\langle\cdot\rangle_{\Omega}$ denotes the mean over the body region. The numerator weights soft-tissue gradients by the corresponding bone gradients; normalising by the mean soft-tissue gradient accounts for differences in overall image smoothness. The body region comprises pixels above $8$ in the $8$-bit base composite, with the full image used when fewer than $1{,}000$ pixels meet this threshold.

The rim ratio focuses on the boundary of the strongest bone regions and compares it with a separate control region:
\begin{equation}
R_{\mathrm{rim}}=\frac{\langle g_{\mathrm{ST}}\rangle_{\Omega_{\mathrm{rim}}}}
{\langle g_{\mathrm{ST}}\rangle_{\Omega_{\mathrm{ctrl}}}+\epsilon_g}.
\label{eq:rim_ratio}
\end{equation}
We define the strong-bone region using bone-projection intensities above the $90$th percentile within the body. Its rim is the difference between two iterations of dilation and erosion with a $3\times3$ square kernel, restricted to the body. The control region comprises body pixels outside eight dilation iterations of the strong-bone region. We calculate the ratio only when both regions contain more than $50$ pixels. Lower values of either ratio indicate less residual bone-shaped structure. Because genuine anatomical edges can coincide with bone, we interpret the ratios comparatively across filling methods rather than treating unity as an absolute target.

\subsection{Statistics and reproducibility}
\label{sec:statistics}
Detector training used fixed training, internal-validation and evaluation partitions. Training seeds varied the initialisation of non-pretrained detector parameters and stochastic training operations, including image sampling/order and augmentation; suppression-model checkpoints, dataset membership and configured hyperparameters were held fixed. TBX11K, Node21 and VinDr-CXR experiments used three detector-training seeds ($42$, $43$ and $44$). Tables~\ref{tab:detection}--\ref{tab:subgroup} and~\ref{tab:arch} report arithmetic means and sample standard deviations across detector runs, describing between-run variability. Checkpoints were selected by internal-validation COCO mAP@[.5:.95]. We average sensitivity across seeds on a common false-positive grid to construct the FROC curves and report the mean of the per-seed CPM values in the legends. CPM differences are summarised descriptively.

Table~\ref{tab:fid} reports population SDs (divisor $n$) over ten size-matched subsampling repetitions; these describe subsampling variability rather than variation across trained generators. Image-level measurements in Tables~\ref{tab:detail} and~\ref{tab:fidelity} are summarised using means and sample SDs (divisor $n-1$) across valid generated images. For the spectral and entropy distances in Table~\ref{tab:detail} and CTR distance in Table~\ref{tab:fidelity}, we report marginal $95\%$ percentile-bootstrap CIs from $10{,}000$ resamples (NumPy random-number seed $20261019$). We resampled CT patients with replacement, retaining all their reconstruction volumes together, and independently resampled the real-reference radiographs. Patient identifiers were derived from the CT-RATE naming convention; each resample used the same patients and real-image draws across generators. We recalculated each distance in every resample, using method-specific valid cases, and took the $2.5$th and $97.5$th percentiles as the interval limits. Estimates remain volume-weighted; these CIs are not simultaneous, multiplicity-adjusted intervals. Table~\ref{tab:abl_boneseg} reports medians and interquartile ranges, while Table~\ref{tab:abl_inpaint} reports between-CT population SDs for both ratios.

For lesion conspicuity, we estimate the mean suppressed-minus-input CNR change and report $95\%$ percentile-bootstrap confidence intervals from $10{,}000$ lesion-level resamples, using $309$ eligible TBX11K TB lesions from $200$ radiographs, $226$ Node21 nodules, $277$ VinDr nodule/mass lesions and $149$ JSRT nodules. For bone removal and detail retention, we compare each baseline with our method using two-sided Wilcoxon signed-rank tests on measurements paired by radiograph, separately for each dataset and evaluation resolution. We report the signed-rank statistic $W$ (the smaller positive or negative rank sum), paired sample size and numerical $p$ value for every comparison in Supplementary Tables~\ref{supp-tab:wilcoxon1024} and~\ref{supp-tab:wilcoxon256}. We use the normal approximation with tie correction, omit zero differences, and apply neither continuity nor multiple-comparison correction. These tests assess whether the distribution of paired differences is symmetric about zero; no degrees-of-freedom parameter is used.

To examine how detection gains relate to bone overlap, we calculate recall at $2$ false positives per image within common overlap groups across methods. These groups contain $46$ overlapped and $185$ clearer Node21 lesions,  $143$ overlapped and $166$ clearer TBX11K lesions, and $398$ overlapped and $2234$ clearer VinDr findings. Bone masks use predicted bone-map intensity $>0.12$ and lesions are assigned to the overlapped group at a bone-mask fraction $\ge15\%$. We further assess class-level associations using Pearson and Spearman correlations across the $14$ VinDr finding classes. For each class, the suppressed-minus-original CPM difference is calculated within each training seed and averaged across the three seeds before correlation; the sample size is 14 classes. We use two-sided tests, allowing associations in either the positive or negative direction. The five suppression methods, two detectors and two correlation measures yield $20$ tests. We report unadjusted $p$ values in Table~\ref{tab:perclass} and use $p<0.05$ as the nominal threshold for these exploratory analyses; no multiple-comparison correction is applied.

\clearpage

\section*{Tables}
\begin{table}[H]
\centering
\caption{Bone-suppression detection on TBX11K, Node21 and VinDr-CXR: FROC CPM and mAP@50 (mean $\pm$ sample SD over seeds 42/43/44; higher is better). Each dataset has two detectors and three input configurations: original ($full$), suppressed-only ($bs$), and combined ($full_{bs}$). Splits and training settings are fixed across methods within each dataset. VinDr training annotations require support from at least two radiologists (min\_conf${=}0.4$). Best mean per column within each dataset is in \textbf{bold}.}
\label{tab:detection}
\small
\setlength{\tabcolsep}{5pt}
\renewcommand{\arraystretch}{1.0}
\begin{adjustbox}{max width=\linewidth}
\begin{tabular}{llcccc}
\toprule
 & & \multicolumn{2}{c}{RetinaNet} & \multicolumn{2}{c}{Faster R-CNN} \\
\cmidrule(lr){3-4}\cmidrule(lr){5-6}
Arm & BS method & CPM\,$\uparrow$ & mAP@50\,$\uparrow$ & CPM\,$\uparrow$ & mAP@50\,$\uparrow$ \\
\midrule
\multicolumn{6}{l}{TBX11K} \\
$full$ & -- & $0.929{\pm}.004$ & $0.752{\pm}.008$ & $0.860{\pm}.011$ & $0.697{\pm}.006$ \\
\cmidrule(lr){1-6}
\multirow{5}{*}{$bs$}
 & \textbf{Ours} & $\mathbf{0.949{\pm}.003}$ & $\mathbf{0.778{\pm}.004}$ & $\mathbf{0.908{\pm}.007}$ & $0.697{\pm}.039$ \\
 & CXR-BS & $0.912{\pm}.012$ & $0.719{\pm}.007$ & $0.896{\pm}.020$ & $0.655{\pm}.009$ \\
 & GL-LCM & $0.921{\pm}.020$ & $0.674{\pm}.058$ & $0.880{\pm}.026$ & $0.652{\pm}.007$ \\
 & DeBoneDiT & $0.916{\pm}.004$ & $0.711{\pm}.035$ & $0.882{\pm}.014$ & $0.656{\pm}.022$ \\
 & BS-LDM & $0.887{\pm}.005$ & $0.640{\pm}.003$ & $0.833{\pm}.061$ & $0.596{\pm}.003$ \\
\cmidrule(lr){1-6}
\multirow{5}{*}{$full_{bs}$}
 & \textbf{Ours} & $0.939{\pm}.007$ & $0.759{\pm}.004$ & $0.883{\pm}.017$ & $\mathbf{0.721{\pm}.026}$ \\
 & CXR-BS & $0.931{\pm}.010$ & $0.753{\pm}.022$ & $0.874{\pm}.017$ & $0.689{\pm}.013$ \\
 & GL-LCM & $0.918{\pm}.002$ & $0.748{\pm}.024$ & $0.894{\pm}.029$ & $0.687{\pm}.027$ \\
 & DeBoneDiT & $0.911{\pm}.005$ & $0.752{\pm}.008$ & $0.879{\pm}.016$ & $0.707{\pm}.010$ \\
 & BS-LDM & $0.917{\pm}.012$ & $0.716{\pm}.025$ & $0.863{\pm}.044$ & $0.688{\pm}.018$ \\
\midrule
\multicolumn{6}{l}{Node21} \\
$full$ & -- & $0.818{\pm}.015$ & $\mathbf{0.629{\pm}.012}$ & $0.821{\pm}.013$ & $0.643{\pm}.008$ \\
\cmidrule(lr){1-6}
\multirow{5}{*}{$bs$}
 & \textbf{Ours} & $0.801{\pm}.014$ & $0.626{\pm}.007$ & $0.822{\pm}.011$ & $0.641{\pm}.030$ \\
 & CXR-BS & $0.714{\pm}.013$ & $0.491{\pm}.014$ & $0.705{\pm}.012$ & $0.545{\pm}.027$ \\
 & GL-LCM & $0.593{\pm}.030$ & $0.378{\pm}.010$ & $0.625{\pm}.012$ & $0.440{\pm}.010$ \\
 & DeBoneDiT & $0.747{\pm}.021$ & $0.557{\pm}.039$ & $0.737{\pm}.020$ & $0.579{\pm}.036$ \\
 & BS-LDM & $0.499{\pm}.038$ & $0.323{\pm}.007$ & $0.553{\pm}.004$ & $0.365{\pm}.027$ \\
\cmidrule(lr){1-6}
\multirow{5}{*}{$full_{bs}$}
 & \textbf{Ours} & $\mathbf{0.823{\pm}.020}$ & $0.619{\pm}.019$ & $\mathbf{0.832{\pm}.004}$ & $\mathbf{0.658{\pm}.026}$ \\
 & CXR-BS & $0.795{\pm}.026$ & $0.606{\pm}.036$ & $0.805{\pm}.004$ & $0.647{\pm}.011$ \\
 & GL-LCM & $0.789{\pm}.043$ & $0.579{\pm}.045$ & $0.807{\pm}.016$ & $0.627{\pm}.015$ \\
 & DeBoneDiT & $0.788{\pm}.019$ & $0.611{\pm}.023$ & $0.801{\pm}.015$ & $0.621{\pm}.015$ \\
 & BS-LDM & $0.805{\pm}.004$ & $0.596{\pm}.012$ & $0.789{\pm}.034$ & $0.605{\pm}.028$ \\
\midrule
\multicolumn{6}{l}{VinDr-CXR} \\
$full$ & -- & $0.349{\pm}.002$ & $0.097{\pm}.001$ & $0.414{\pm}.038$ & $0.113{\pm}.008$ \\
\cmidrule(lr){1-6}
\multirow{5}{*}{$bs$} & \textbf{Ours} & $0.357{\pm}.017$ & $0.098{\pm}.004$ & $0.400{\pm}.014$ & $0.109{\pm}.003$ \\
 & CXR-BS & $0.331{\pm}.029$ & $0.092{\pm}.008$ & $0.334{\pm}.011$ & $0.101{\pm}.001$ \\
 & GL-LCM & $0.320{\pm}.010$ & $0.086{\pm}.002$ & $0.371{\pm}.029$ & $0.102{\pm}.009$ \\
 & DeBoneDiT & $0.331{\pm}.020$ & $0.093{\pm}.007$ & $0.363{\pm}.018$ & $0.107{\pm}.006$ \\
 & BS-LDM & $0.300{\pm}.020$ & $0.079{\pm}.005$ & $0.398{\pm}.033$ & $0.102{\pm}.001$ \\
\cmidrule(lr){1-6}
\multirow{5}{*}{$full_{bs}$} & \textbf{Ours} & $\mathbf{0.379{\pm}.018}$ & $\mathbf{0.102{\pm}.004}$ & $\mathbf{0.431{\pm}.060}$ & $\mathbf{0.120{\pm}.016}$ \\
 & CXR-BS & $0.352{\pm}.023$ & $0.094{\pm}.004$ & $0.408{\pm}.019$ & $0.113{\pm}.009$ \\
 & GL-LCM & $0.362{\pm}.010$ & $0.099{\pm}.003$ & $0.403{\pm}.028$ & $0.114{\pm}.004$ \\
 & DeBoneDiT & $0.334{\pm}.021$ & $0.088{\pm}.013$ & $0.396{\pm}.054$ & $0.112{\pm}.010$ \\
 & BS-LDM & $0.339{\pm}.018$ & $0.095{\pm}.003$ & $0.391{\pm}.056$ & $0.111{\pm}.010$ \\
\bottomrule
\end{tabular}
\end{adjustbox}
\end{table}

\begin{table}[H]
\centering
\caption{Exploratory bone-overlap analysis for our suppression method: recall at 2 false positives per image (three-seed mean $\pm$ sample SD). $\Delta$ is the mean seed-matched change from the same detector's $full$ baseline. Bold positive $\Delta_{\mathrm{ov}}-\Delta_{\mathrm{clear}}$ values indicate a greater gain for overlapped lesions. Subgroup sizes ($\ge15\%$ / $<15\%$ bone overlap): TBX11K $143/166$; Node21 $46/185$; VinDr-CXR $398/2234$ across all 14 findings. Comparisons with other suppression methods are in Supplementary Tables~\ref{supp-tab:overlap_tbx11k}--\ref{supp-tab:overlap_vindr}.}
\label{tab:subgroup}
\footnotesize
\setlength{\tabcolsep}{3pt}
\begin{adjustbox}{max width=\linewidth}
\begin{tabular}{llccccc}
\toprule
Dataset / Detector & Arm & Overlap recall & $\Delta_{\mathrm{ov}}$ & Clearer recall & $\Delta_{\mathrm{clear}}$ & $\Delta_{\mathrm{ov}}-\Delta_{\mathrm{clear}}$ \\
\midrule
TBX11K / RetinaNet & $full$ & $0.979\pm0.007$ & -- & $0.934\pm0.022$ & -- & -- \\
 & $bs$ & $0.988\pm0.008$ & $+0.009$ & $0.950\pm0.009$ & $+0.016$ & $-0.007$ \\
 & $full_{bs}$ & $0.970\pm0.008$ & $-0.009$ & $0.940\pm0.028$ & $+0.006$ & $-0.015$ \\
\midrule
TBX11K / Faster R-CNN & $full$ & $0.890\pm0.029$ & -- & $0.835\pm0.003$ & -- & -- \\
 & $bs$ & $0.935\pm0.016$ & $+0.044$ & $0.894\pm0.009$ & $+0.058$ & $-0.014$ \\
 & $full_{bs}$ & $0.921\pm0.016$ & $+0.030$ & $0.851\pm0.030$ & $+0.016$ & $\mathbf{+0.014}$ \\
\midrule
Node21 / RetinaNet & $full$ & $0.826\pm0.022$ & -- & $0.874\pm0.003$ & -- & -- \\
 & $bs$ & $0.841\pm0.013$ & $+0.014$ & $0.859\pm0.028$ & $-0.014$ & $\mathbf{+0.029}$ \\
 & $full_{bs}$ & $0.855\pm0.025$ & $+0.029$ & $0.865\pm0.024$ & $-0.009$ & $\mathbf{+0.038}$ \\
\midrule
Node21 / Faster R-CNN & $full$ & $0.841\pm0.050$ & -- & $0.903\pm0.025$ & -- & -- \\
 & $bs$ & $0.870\pm0.022$ & $+0.029$ & $0.901\pm0.019$ & $-0.002$ & $\mathbf{+0.031}$ \\
 & $full_{bs}$ & $0.877\pm0.013$ & $+0.036$ & $0.895\pm0.011$ & $-0.007$ & $\mathbf{+0.043}$ \\
\midrule
VinDr-CXR / RetinaNet & $full$ & $0.360\pm0.017$ & -- & $0.438\pm0.003$ & -- & -- \\
 & $bs$ & $0.373\pm0.030$ & $+0.013$ & $0.442\pm0.028$ & $+0.004$ & $\mathbf{+0.009}$ \\
 & $full_{bs}$ & $0.381\pm0.004$ & $+0.021$ & $0.463\pm0.012$ & $+0.024$ & $-0.003$ \\
\midrule
VinDr-CXR / Faster R-CNN & $full$ & $0.386\pm0.079$ & -- & $0.501\pm0.081$ & -- & -- \\
 & $bs$ & $0.338\pm0.037$ & $-0.049$ & $0.469\pm0.015$ & $-0.032$ & $-0.016$ \\
 & $full_{bs}$ & $0.432\pm0.127$ & $+0.046$ & $0.512\pm0.101$ & $+0.011$ & $\mathbf{+0.035}$ \\
\bottomrule
\end{tabular}
\end{adjustbox}
\end{table}

\begin{table}[H]
\centering
\caption{Class-level correlations between bone-overlap prevalence and detection change ($\Delta$CPM $=bs-full$) across VinDr's 14 classes ($n=14$). For each class, seed-matched CPM changes are averaged over seeds 42, 43 and 44 before calculating Pearson and Spearman coefficients. Parentheses give two-sided, unadjusted $p$ values. These exploratory correlations describe associations, not differences between methods.}
\label{tab:perclass}
\footnotesize
\setlength{\tabcolsep}{3pt}
\begin{adjustbox}{max width=\linewidth}
\begin{tabular}{llcc}
\toprule
Method & Detector & Pearson $r$ ($p$) & Spearman $\rho$ ($p$) \\
\midrule
Ours & RetinaNet & $+0.411$ (0.144) & $+0.398$ (0.159) \\
CXR-BS & RetinaNet & $-0.155$ (0.598) & $-0.077$ (0.794) \\
GL-LCM & RetinaNet & $-0.004$ (0.988) & $+0.007$ (0.982) \\
DeBoneDiT & RetinaNet & $-0.239$ (0.410) & $+0.020$ (0.946) \\
BS-LDM & RetinaNet & $-0.241$ (0.407) & $-0.095$ (0.748) \\
Ours & Faster R-CNN & $+0.282$ (0.329) & $+0.332$ (0.246) \\
CXR-BS & Faster R-CNN & $-0.352$ (0.217) & $-0.349$ (0.221) \\
GL-LCM & Faster R-CNN & $-0.197$ (0.499) & $-0.015$ (0.958) \\
DeBoneDiT & Faster R-CNN & $-0.139$ (0.635) & $+0.077$ (0.794) \\
BS-LDM & Faster R-CNN & $-0.183$ (0.531) & $-0.367$ (0.197) \\
\bottomrule
\end{tabular}
\end{adjustbox}
\end{table}

\begin{table}[H]
\centering
\caption{Bone-overlap prevalence per dataset (a lesion is ``overlapped'' if $\ge 15\%$ of its box pixels are bone). Bone-suppression benefit in Sec.~\ref{sec:bs_results} is clearest where this prevalence is highest (TBX11K).}
\label{tab:prevalence}
\small
\begin{adjustbox}{max width=\linewidth}
\begin{tabular}{lccc}
\toprule
Dataset & Lesions & Overlapped ($\ge 15\%$) & No bone contact \\
\midrule
TBX11K & 309 & 143 \ (46.3\%) & 21 \ (6.8\%) \\
Node21 & 231 & 46 \ (19.9\%) & 98 \ (42.4\%) \\
VinDr-CXR & 2{,}632 & 398 \ (15.1\%) & 1{,}373 \ (52.2\%) \\
\bottomrule
\end{tabular}
\end{adjustbox}
\end{table}

\begin{table}[H]
\centering
\caption{Lesion conspicuity change $\Delta$CNR (mean [$95\%$ percentile-bootstrap CI]) on TBX11K TB lesions ($n{=}309$ from $200$ radiographs), Node21 nodules ($n{=}226$ of $231$), the VinDr nodule/mass class ($n{=}277$ of $286$), and JSRT nodules ($n{=}149$; boxes from the published clinical coordinates), using every lesion whose box lies at least $25\%$ inside the (dilated) rib$\cup$clavicle region of the CXAS segmenter~\cite{cxas}. The selection is made once per lesion and is identical for every method; each method is analysed at the smaller of the dataset resolution and its native output resolution (CXR-BS at $256 \times 256$). Positive $=$ lesion more conspicuous after suppression. \textbf{Bold} $=$ CI entirely above zero (on JSRT, the only CI containing zero); $^{\dagger}$ $=$ CI entirely below zero.}
\label{tab:cnr}
\small
\setlength{\tabcolsep}{6pt}
\renewcommand{\arraystretch}{1.15}
\begin{adjustbox}{max width=\linewidth}
\begin{tabular}{lcccc}
\toprule
Method & TBX11K & Node21 & VinDr nod./mass & JSRT \\
 & $n=309$ & $n=226$ & $n=277$ & $n=149$ \\
\midrule
\textbf{Ours}
 & \shortstack{$\mathbf{+0.047}$\\$[+0.038,+0.055]$}
 & \shortstack{$\mathbf{+0.067}$\\$[+0.025,+0.109]$}
 & \shortstack{$\mathbf{+0.049}$\\$[+0.022,+0.076]$}
 & \shortstack{$\mathbf{+0.008}$\\$[-0.022,+0.039]$} \\
\addlinespace[5pt]
CXR-BS
 & \shortstack{$\mathbf{+0.055}$\\$[+0.048,+0.062]$}
 & \shortstack{$+0.005$\\$[-0.022,+0.034]$}
 & \shortstack{$-0.014$\\$[-0.036,+0.007]$}
 & \shortstack{$-0.050^{\dagger}$\\$[-0.083,-0.018]$} \\
\addlinespace[5pt]
GL-LCM
 & \shortstack{$\mathbf{+0.042}$\\$[+0.035,+0.048]$}
 & \shortstack{$-0.091^{\dagger}$\\$[-0.123,-0.061]$}
 & \shortstack{$-0.055^{\dagger}$\\$[-0.079,-0.034]$}
 & \shortstack{$-0.156^{\dagger}$\\$[-0.190,-0.124]$} \\
\addlinespace[5pt]
DeBoneDiT
 & \shortstack{$\mathbf{+0.023}$\\$[+0.018,+0.028]$}
 & \shortstack{$-0.030^{\dagger}$\\$[-0.045,-0.015]$}
 & \shortstack{$-0.030^{\dagger}$\\$[-0.044,-0.017]$}
 & \shortstack{$-0.129^{\dagger}$\\$[-0.159,-0.101]$} \\
\addlinespace[5pt]
BS-LDM
 & \shortstack{$\mathbf{+0.044}$\\$[+0.038,+0.050]$}
 & \shortstack{$-0.102^{\dagger}$\\$[-0.129,-0.076]$}
 & \shortstack{$-0.047^{\dagger}$\\$[-0.068,-0.029]$}
 & \shortstack{$-0.139^{\dagger}$\\$[-0.171,-0.109]$} \\
\bottomrule
\end{tabular}
\end{adjustbox}
\end{table}

\begin{table}[H]
\centering
\caption{Bone-removal completeness vs.\ clear-lung detail retention (mean$\pm$SD); the retention target is $\approx\!1$. Upper block: the $1024 \times 1024$-native methods at $1024 \times 1024$ (paired Wilcoxon vs.\ ours). Lower block: CXR-BS on its native $256 \times 256$ grid, with our output resampled to the same grid (paired Wilcoxon vs.\ ours); the full $256 \times 256$ comparison of all methods is Supplementary Table~\ref{supp-tab:supp_antiblur256}. \textbf{Bold} $=$ retention closest to $1$ per dataset within the $1024 \times 1024$ block; removal is left unbolded because lost detail inflates it (see text). $n{=}350$ (Node21), $3000$ (VinDr), $247$ (JSRT). Signed-rank statistics and numerical $p$ values are given in Supplementary Tables~\ref{supp-tab:wilcoxon1024} and~\ref{supp-tab:wilcoxon256}.}
\label{tab:antiblur}
\small
\setlength{\tabcolsep}{3pt}
\begin{adjustbox}{max width=\linewidth}
\begin{tabular}{lccccccc}
\toprule
 & Eval. & \multicolumn{2}{c}{Node21} & \multicolumn{2}{c}{VinDr} & \multicolumn{2}{c}{JSRT} \\
\cmidrule(lr){3-4}\cmidrule(lr){5-6}\cmidrule(lr){7-8}
Method & res & removal & retention & removal & retention & removal & retention \\
\midrule
\textbf{Ours} & $1024$ & $0.32{\pm}.08$ & $\mathbf{1.12{\pm}.24}$ & $0.35{\pm}.08$ & $\mathbf{1.19{\pm}.19}$ & $0.25{\pm}.05$ & $\mathbf{1.05{\pm}.06}$ \\
GL-LCM        & $1024$ & $0.58{\pm}.04$ & $0.48{\pm}.22$ & $0.56{\pm}.06$ & $1.45{\pm}1.27$ & $0.44{\pm}.05$ & $1.61{\pm}.45$ \\
DeBoneDiT     & $1024$ & $0.34{\pm}.06$ & $0.55{\pm}.27$ & $0.47{\pm}.08$ & $1.23{\pm}1.01$ & $0.40{\pm}.05$ & $1.27{\pm}.39$ \\
BS-LDM        & $1024$ & $0.54{\pm}.08$ & $0.36{\pm}.20$ & $0.47{\pm}.12$ & $1.37{\pm}1.46$ & $0.42{\pm}.05$ & $1.57{\pm}.52$ \\
\midrule
CXR-BS (native grid) & $256$ & $0.32{\pm}.03$ & $0.84{\pm}.16$ & $0.31{\pm}.06$ & $0.91{\pm}.20$ & $0.16{\pm}.03$ & $1.24{\pm}.07$ \\
\textbf{Ours} (resampled)     & $256$ & $0.31{\pm}.08$ & $0.92{\pm}.22$ & $0.34{\pm}.08$ & $0.90{\pm}.20$ & $0.25{\pm}.05$ & $0.88{\pm}.10$ \\
\bottomrule
\end{tabular}
\end{adjustbox}
\end{table}

\begin{table}[H]
\centering
\caption{Distributional realism to real CXR in a CXR-pretrained feature space (lower is better), reported as mean $\pm$ population SD over ten size-matched subsampling repetitions. All methods use the same eligible pool of $22{,}124$ CT-RATE volumes. Each repetition compares $18{,}000$ images per distribution against VinDr or $22{,}124$ against CheXpert; the larger pool is subsampled without replacement and the smaller retained in full. Both KID means and SDs are scaled by $10^3$. Our two arms are \base{} (additive sum of the component projections, Sec.~\ref{sec:compositing}) and the translated radiograph \trans{} (Stage~3). Best in \textbf{bold}.}
\label{tab:fid}
\footnotesize
\setlength{\tabcolsep}{3pt}
\renewcommand{\arraystretch}{1.2}
\begin{tabular*}{\linewidth}{@{}l@{\extracolsep{\fill}}cccc@{}}
\toprule
Generator & VinDr FID & VinDr KID & CheXpert FID & CheXpert KID \\
\midrule
\textbf{Ours} (\base{}) & $14.796 \pm 0.013$ & $30.885 \pm 0.037$ & $23.633 \pm 0.150$ & $53.086 \pm 0.110$ \\
\textbf{Ours} (\trans{}) & $\mathbf{8.173 \pm 0.008}$ & $\mathbf{13.886 \pm 0.015}$ & $\mathbf{16.270 \pm 0.111}$ & $\mathbf{32.020 \pm 0.100}$ \\
DeepDRR & $14.996 \pm 0.019$ & $30.778 \pm 0.053$ & $27.871 \pm 0.094$ & $60.357 \pm 0.114$ \\
Plastimatch & $14.506 \pm 0.016$ & $29.567 \pm 0.037$ & $28.512 \pm 0.151$ & $61.543 \pm 0.107$ \\
nanoDRR/DiffDRR & $14.936 \pm 0.026$ & $30.846 \pm 0.072$ & $28.725 \pm 0.110$ & $62.644 \pm 0.154$ \\
TIGRE & $16.454 \pm 0.013$ & $34.794 \pm 0.031$ & $29.737 \pm 0.204$ & $65.090 \pm 0.089$ \\
\bottomrule
\end{tabular*}
\end{table}

\begin{table}[H]
\centering
\caption{Fine-scale detail and texture over the same two arms as Table~\ref{tab:fid}, evaluated at $512\times512$ pixels. Panel A: spectral $L_2$ distance between mean radially averaged power spectra and local-entropy Wasserstein distance to real VinDr images (lower is better), with marginal $95\%$ bootstrap CIs and valid-volume counts for entropy ($n_{\rm entropy}$). Panel B: whole-image and lung-field Laplacian variance (mean $\pm$ sample SD across valid images; higher means sharper), with valid-volume counts for lung sharpness ($n_{\rm lung}$). The common pool contains $22{,}124$ volumes from $14{,}538$ patients; all volumes contribute to whole-image sharpness and spectral distance. Real-reference counts are $3{,}000$ for spectra/sharpness and $2{,}995$ for entropy. Real VinDr whole-image/lung sharpness is $281.9\pm155.2$/$290.8\pm183.8$. Best synthetic values in \textbf{bold} denote descriptive rankings.}
\label{tab:detail}
\footnotesize
\setlength{\tabcolsep}{3pt}
\renewcommand{\arraystretch}{1.2}
\textbf{A. Distribution distances}\par\smallskip
\begin{tabular*}{\linewidth}{@{}l@{\extracolsep{\fill}}ccccc@{}}
\toprule
Generator & Spec.\ $L_2$ & $95\%$ CI & Entropy W & $95\%$ CI & $n_{\rm entropy}$ \\
\midrule
\textbf{Ours} (\base{}) & $19.73$ & $[19.50, 19.96]$ & $0.403$ & $[0.392, 0.415]$ & 22{,}115 \\
\textbf{Ours} (\trans{}) & $\mathbf{2.27}$ & $[2.24, 2.30]$ & $\mathbf{0.309}$ & $[0.298, 0.320]$ & 22{,}101 \\
DeepDRR & $15.75$ & $[15.61, 15.88]$ & $0.749$ & $[0.729, 0.769]$ & 18{,}711 \\
Plastimatch & $19.01$ & $[18.84, 19.18]$ & $0.824$ & $[0.808, 0.841]$ & 21{,}898 \\
nanoDRR/DiffDRR & $19.54$ & $[19.38, 19.70]$ & $0.833$ & $[0.817, 0.850]$ & 21{,}886 \\
TIGRE & $17.49$ & $[17.35, 17.62]$ & $0.955$ & $[0.937, 0.973]$ & 21{,}756 \\
\bottomrule
\end{tabular*}
\par\vspace{8pt}\textbf{B. Image-level sharpness and valid counts}\par\smallskip
\begin{tabular*}{\linewidth}{@{}l@{\extracolsep{\fill}}ccc@{}}
\toprule
Generator & Whole-image var. & Lung var. & $n_{\rm lung}$ \\
\midrule
\textbf{Ours} (\base{}) & $16.9 \pm 23.4$ & $18.1 \pm 22.7$ & 22{,}006 \\
\textbf{Ours} (\trans{}) & $\mathbf{31.0 \pm 25.7}$ & $\mathbf{23.6 \pm 23.3}$ & 22{,}051 \\
DeepDRR & $28.1 \pm 27.9$ & $9.2 \pm 4.3$ & 21{,}632 \\
Plastimatch & $14.0 \pm 19.1$ & $8.5 \pm 9.1$ & 21{,}645 \\
nanoDRR/DiffDRR & $12.7 \pm 17.3$ & $6.7 \pm 4.1$ & 21{,}659 \\
TIGRE & $18.9 \pm 20.1$ & $6.2 \pm 4.4$ & 21{,}660 \\
\bottomrule
\end{tabular*}
\end{table}

\begin{table}[H]
\centering
\caption{Anatomical fidelity over the same two arms as Table~\ref{tab:fid}. Panel A: transfer-probe AUROC against source-CT labels for a frozen classifier trained only on real CheXpert radiographs; AUROC-4 is the unweighted mean across four findings. All methods use $22{,}124$ reconstruction volumes from $14{,}538$ patients. Positive/negative volume counts are $2{,}577/19{,}547$ (effusion), $5{,}656/16{,}468$ (atelectasis), $2{,}656/19{,}468$ (cardiomegaly) and $4{,}073/18{,}051$ (consolidation). Panel B: structures detected (/14) and cross-modal lung IoU are mean $\pm$ sample SD across images. Structure count uses all volumes; IoU uses $21{,}578$ volumes from $14{,}524$ patients whose Stage-1 lung segmentation occupies at least $1\%$ of the padded CT volume and compares the generated-image lung mask with the projected CT bounding-box region. CTR dist. is the Wasserstein distance between generated and real cardiothoracic-ratio distributions, with a marginal $95\%$ bootstrap CI; $n_{\rm CTR}$ gives valid generated-volume counts and the real reference contains $2{,}999$ images. Higher is better except CTR dist.; best point estimates per column in \textbf{bold}.}
\label{tab:fidelity}
\footnotesize
\setlength{\tabcolsep}{3pt}
\renewcommand{\arraystretch}{1.2}
\textbf{A. Transfer-probe AUROC}\par\smallskip
\begin{tabular*}{\linewidth}{@{}l@{\extracolsep{\fill}}ccccc@{}}
\toprule
Generator & Effusion & Atelectasis & Cardiomegaly & Consolidation & AUROC-4 \\
\midrule
\textbf{Ours} (\base{}) & $0.850$ & $0.630$ & $0.818$ & $\mathbf{0.773}$ & $0.768$ \\
\textbf{Ours} (\trans{}) & $0.861$ & $0.647$ & $0.827$ & $0.759$ & $0.774$ \\
DeepDRR & $\mathbf{0.885}$ & $\mathbf{0.669}$ & $\mathbf{0.874}$ & $\mathbf{0.773}$ & $\mathbf{0.800}$ \\
Plastimatch & $0.848$ & $0.646$ & $0.857$ & $0.749$ & $0.775$ \\
nanoDRR/DiffDRR & $0.842$ & $0.643$ & $0.854$ & $0.747$ & $0.772$ \\
TIGRE & $0.861$ & $0.648$ & $0.865$ & $0.754$ & $0.782$ \\
\bottomrule
\end{tabular*}
\par\vspace{8pt}\textbf{B. Anatomical measurements}\par\smallskip
\begin{tabular*}{\linewidth}{@{}l@{\extracolsep{\fill}}ccccc@{}}
\toprule
Generator & Structures /14 & CTR dist. & $95\%$ CI & $n_{\rm CTR}$ & Lung IoU \\
\midrule
\textbf{Ours} (\base{}) & $12.95 \pm 2.18$ & $0.078$ & $[0.076, 0.080]$ & 21{,}454 & $0.449 \pm 0.100$ \\
\textbf{Ours} (\trans{}) & $\mathbf{13.42 \pm 1.97}$ & $\mathbf{0.025}$ & $[0.023, 0.027]$ & 21{,}499 & $\mathbf{0.457 \pm 0.102}$ \\
DeepDRR & $13.02 \pm 2.33$ & $0.067$ & $[0.065, 0.069]$ & 21{,}253 & $0.408 \pm 0.092$ \\
Plastimatch & $12.58 \pm 2.34$ & $0.056$ & $[0.054, 0.058]$ & 21{,}166 & $0.416 \pm 0.093$ \\
nanoDRR/DiffDRR & $12.51 \pm 2.34$ & $0.059$ & $[0.057, 0.061]$ & 21{,}140 & $0.417 \pm 0.093$ \\
TIGRE & $12.75 \pm 2.33$ & $0.088$ & $[0.085, 0.090]$ & 21{,}098 & $0.414 \pm 0.093$ \\
\bottomrule
\end{tabular*}
\end{table}

\begin{table}[H]
\centering
\caption{Bone-mask construction on the fixed $5{,}000$-CT subset; stages are cumulative. Components, volume and Largest-CC are reported as median [IQR] across CTs. Components: 3D connected components per mask (the mean is dominated by thin-slice outliers with up to $\sim$$1.4$\,M speckle components). Largest-CC: fraction of mask voxels in the single largest component. Pruning removes a median $99.2\%$ of components. Volume retention ($99.31\%$) is calculated for each CT as post-pruning volume divided by pre-pruning volume, then summarised by its median. Dividing the separately reported median volumes, $1236/1248$, instead gives $99.0\%$; these are different summaries.}
\label{tab:abl_boneseg}
\footnotesize
\setlength{\tabcolsep}{3pt}
\begin{adjustbox}{max width=\linewidth}
\begin{tabular}{lccc}
\toprule
Bone mask & Components $\downarrow$ & Volume (cm$^3$) & Largest-CC \\
\midrule
(a) threshold ($200$\,HU) & 318 [198--495] & 932 [793--1108] & 0.856 [0.763--0.956] \\
(b) $+$ iterative ladder & 201 [127--337] & 1248 [1066--1505] & 0.975 [0.892--0.991] \\
(c) $+$ CC pruning (ours) & \textbf{2 [1--3]} & 1236 [1055--1494] & \textbf{0.986 [0.901--1.000]} \\
\bottomrule
\end{tabular}
\end{adjustbox}
\end{table}

\begin{table}[H]
\centering
\caption{Soft-tissue void-fill ablation on the same $5{,}000$ CTs (mean $\pm$ population SD across CTs; $n=5{,}000$ valid measurements per method for both ratios). Bone-edge ratio: bone-gradient-weighted soft-tissue gradient magnitude, normalised by its mean over the body (Eq.~\eqref{eq:bone_edge_ratio}). Rim ratio: mean soft-tissue gradient magnitude on the rim of the strongest bone regions relative to a separate control region (Eq.~\eqref{eq:rim_ratio}). Lower values indicate less bone-shaped residue; genuine anatomical edges can also contribute. Best per column in \textbf{bold}.}
\label{tab:abl_inpaint}
\footnotesize
\setlength{\tabcolsep}{3pt}
\begin{adjustbox}{max width=\linewidth}
\begin{tabular}{lccc}
\toprule
Fill & Bone-edge ratio $\downarrow$ & $\Delta$ vs.\ best & Rim ratio $\downarrow$ \\
\midrule
\textbf{Diffusion (ours)} & $\mathbf{1.1455 \pm 0.069}$ & -- & $0.992 \pm 0.131$ \\
Telea (slice-wise) & $1.1579 \pm 0.081$ & $+0.012$ & $0.995 \pm 0.137$ \\
Nearest value & $1.1615 \pm 0.080$ & $+0.016$ & $0.995 \pm 0.135$ \\
Constant $-50$\,HU & $1.1618 \pm 0.076$ & $+0.016$ & $\mathbf{0.990 \pm 0.133}$ \\
Water ($0$\,HU) & $1.1966 \pm 0.078$ & $+0.051$ & $1.046 \pm 0.140$ \\
Air ($-1000$\,HU) & $1.3049 \pm 0.063$ & $+0.159$ & $1.320 \pm 0.116$ \\
\bottomrule
\end{tabular}
\end{adjustbox}
\end{table}

\begin{table}[H]
\centering
\caption{Suppression-model backbone ablation, FROC CPM on the $bs$ arm (mean $\pm$ sample SD across three detector-training seeds, 42/43/44, using fixed suppression-model checkpoints). FlowMatch (flow-matching DiT-B/2) is trained at $512 \times 512$; ours at $1024 \times 1024$.}
\label{tab:arch}
\footnotesize
\setlength{\tabcolsep}{3pt}
\begin{adjustbox}{max width=\linewidth}
\begin{tabular}{llcc}
\toprule
Dataset & Detector & Ours (SwinV2$+$UNet++)\,$\uparrow$ & FlowMatch\,$\uparrow$ \\
\midrule
\multirow{2}{*}{TBX11K}
 & RetinaNet    & $\mathbf{0.949 \pm 0.003}$ & $0.907 \pm 0.012$ \\
 & Faster R-CNN & $\mathbf{0.908 \pm 0.007}$ & $0.892 \pm 0.007$ \\
\midrule
\multirow{2}{*}{Node21}
 & RetinaNet    & $\mathbf{0.801 \pm 0.014}$ & $0.573 \pm 0.013$ \\
 & Faster R-CNN & $\mathbf{0.822 \pm 0.011}$ & $0.603 \pm 0.014$ \\
\bottomrule
\end{tabular}
\end{adjustbox}
\end{table}

\begin{table}[H]
\centering
\caption{Inference cost on a single NVIDIA L40S (48\,GB). Our single-pass model runs at higher resolution and $6\text{--}110\times$ higher throughput than the iterative flow-matching model, whose cost scales linearly with the number of ODE steps.}
\label{tab:runtime}
\footnotesize
\setlength{\tabcolsep}{3pt}
\begin{adjustbox}{max width=\linewidth}
\begin{tabular}{lccccc}
\toprule
Model & Params & Res & Steps & Time/img & Thpt.\ (img/s) \\
\midrule
SwinV2-B$+$UNet++ (ours) & $96$M & $1024 \times 1024$ & 1 & $0.099$\,s & $\mathbf{10.1}$ \\
\midrule
\multirow{3}{*}{FlowMatch}
 & \multirow{3}{*}{$214$M} & \multirow{3}{*}{$512 \times 512$} & 50 & $0.629$\,s & $1.6$ \\
 & & & 100 & $1.17$\,s & $0.9$ \\
 & & & 1000 & $10.9$\,s & $0.1$ \\
\bottomrule
\end{tabular}
\end{adjustbox}
\end{table}

\begin{table}[H]
\centering
\caption{Elemental composition (mass fractions) assigned to each anatomical component, based on ICRU Report~44 reference tissue compositions~\cite{icru44}; the listed fractions enter the mixture rule, which normalises them to unit sum. The energy-dependent mass attenuation coefficient $\mu_c(E)$ is computed from these compositions via the NIST XCOM mixture rule~\cite{xcom}. Dashes denote elements absent from that material.}
\label{tab:materials}
\small
\setlength{\tabcolsep}{5pt}
\renewcommand{\arraystretch}{1.2}
\begin{adjustbox}{max width=\linewidth}
\begin{tabular}{lcccccccc}
\toprule
Component & H & C & N & O & Mg & P & Ca & Fe \\
\midrule
Lung        & 0.103 & 0.105 & 0.031 & 0.749 & --    & 0.002    & --    & 0.001 \\
Soft tissue & 0.102 & 0.143 & 0.034 & 0.708 & --    & 0.003    & --    & --    \\
Bone        & 0.034 & 0.155 & 0.042 & 0.435 & 0.002 & 0.103 & 0.225 & --    \\
\bottomrule
\end{tabular}
\end{adjustbox}
\end{table}

\begin{table}[H]
\centering
\caption{Training configuration of the two suppression models.}
\label{tab:supp_bs_config}
\small
\begin{tabular}{lcc}
\toprule
 & Bone suppression & Lung-component suppression \\
\midrule
Predicted component & bone $P_{\mathrm{BN}}$ & lung/vascular $P_{\mathrm{LN}}$ \\
Optimizer & \multicolumn{2}{c}{AdamW, weight decay $0.01$} \\
Peak learning rate & $5{\times}10^{-5}$ & $1{\times}10^{-4}$ \\
Schedule & \multicolumn{2}{c}{one-cycle cosine, $10\%$ warm-up (div.\ factors $250/100$)} \\
Batch size (per GPU $\times$ accum.\ $\times$ GPUs) & $8\times4\times4$ & $14\times6\times4$ \\
Effective batch & $128$ & $336$ \\
Epochs $\times$ samples/epoch & \multicolumn{2}{c}{$500\times64{,}000$} \\
Input resolution & \multicolumn{2}{c}{$1024^2$} \\
Precision / grad.\ clip & \multicolumn{2}{c}{fp16 / $1.0$ (norm)} \\
\bottomrule
\end{tabular}
\end{table}

\begin{table}[H]
\centering
\caption{Detector training recipe, held fixed across all arms and methods.}
\label{tab:detcfg}
\footnotesize
\setlength{\tabcolsep}{4pt}
\begin{adjustbox}{max width=\linewidth}
\begin{tabular}{ll}
\toprule
Item & Value \\
\midrule
Detectors & torchvision RetinaNet-R50-FPN-v2 and Faster R-CNN-R50-FPN-v2 \\
Initialisation & ImageNet backbone, detection head from scratch. \emph{Exception:} VinDr-CXR loads the \\
 & full COCO-pretrained detector with the class head re-initialised, applied identically \\
 & to all VinDr arms \\
Optimiser & AdamW, weight decay $10^{-4}$ \\
Learning rate & $2.5\times10^{-4}$, cosine annealing with a $3$-epoch linear warm-up; Behrendt's base \\
 & $10^{-4}$ at batch $16$, $\sqrt{\cdot}$-scaled to our effective batch $96$ \\
Effective batch & $96$ for both detectors (RetinaNet $24\times4$ GPUs; Faster R-CNN $8\times4$ GPUs with \\
 & gradient accumulation $3$, its heavier ROI head precluding a larger per-GPU batch) \\
Resolution & TBX11K $512^2$; Node21 and VinDr-CXR $1024^2$. The detector transform is pinned to \\
 & the square size, with no internal re-resize \\
Epochs & TBX11K $50$, Node21 $50$, VinDr-CXR $60$ \\
Losses / anchors & RetinaNet focal loss ($\alpha{=}0.25$, $\gamma{=}2.0$); default FPN anchors, not hand-tuned \\
Two-channel input & the $full_{bs}$ arm stacks $full$ and $bs$; the first convolution is inflated by \\
 & channel-mean replication, leaving the pretrained weights otherwise unchanged \\
Model selection & best COCO mAP@[.5:.95] on an internal validation split carved from the training pool \\ & (TBX11K: $990$ images held out of the official training set; the official validation set is \\ & used only for evaluation); the evaluation split is scored at that checkpoint \\
Repeats & seeds $42/43/44$, mean$\pm$SD on a frozen per-dataset split \\
\bottomrule
\end{tabular}
\end{adjustbox}
\end{table}

\clearpage
\setcounter{section}{0}\setcounter{subsection}{0}\setcounter{table}{0}\setcounter{figure}{0}\setcounter{equation}{0}
\renewcommand{\thesection}{S\arabic{section}}
\renewcommand{\theHsection}{supp.\arabic{section}}
\renewcommand{\thetable}{S\arabic{table}}
\renewcommand{\thefigure}{S\arabic{figure}}
\renewcommand{\theHtable}{supp.\arabic{table}}
\renewcommand{\theHfigure}{supp.\arabic{figure}}
\renewcommand{\figurename}{Supplementary Figure}
\renewcommand{\tablename}{Supplementary Table}
\begin{center}
{\Large\bfseries Supplementary Information\par}
\end{center}
\noindent This file contains additional results and illustrations. All methodological descriptions and training configurations are provided in the main manuscript.
\section{BS-Diff detection results}
\label{supp-sec:bsdiff}
Table~\ref{supp-tab:bsdiff} reports the detection results for BS-Diff, evaluated under the
identical protocol as the methods in Table~\mainref{tab:detection}{1} of the main manuscript (same detector recipes,
splits, and seeds). BS-Diff is the predecessor of the three diffusion-based systems compared in the
main text, which represent that lineage in the main-text tables; its results are therefore reported here. The $bs$ arm is the weakest of all methods on TBX11K and
fails to train on Node21 (validation mAP flat at ${\approx}0.01$ for the full
schedule in all six runs); the $full_{bs}$ arm recovers through the intact
full-radiograph channel. Only CPM is reported for BS-Diff because the mAP@50 logs of these runs were not retained.

\section{Soft-tissue void-fill: qualitative comparison}
\label{supp-ssec:inpaint_fig}
\begin{figure}[H]
\centering
\vspace{6mm}
\begin{overpic}[width=160.0mm]{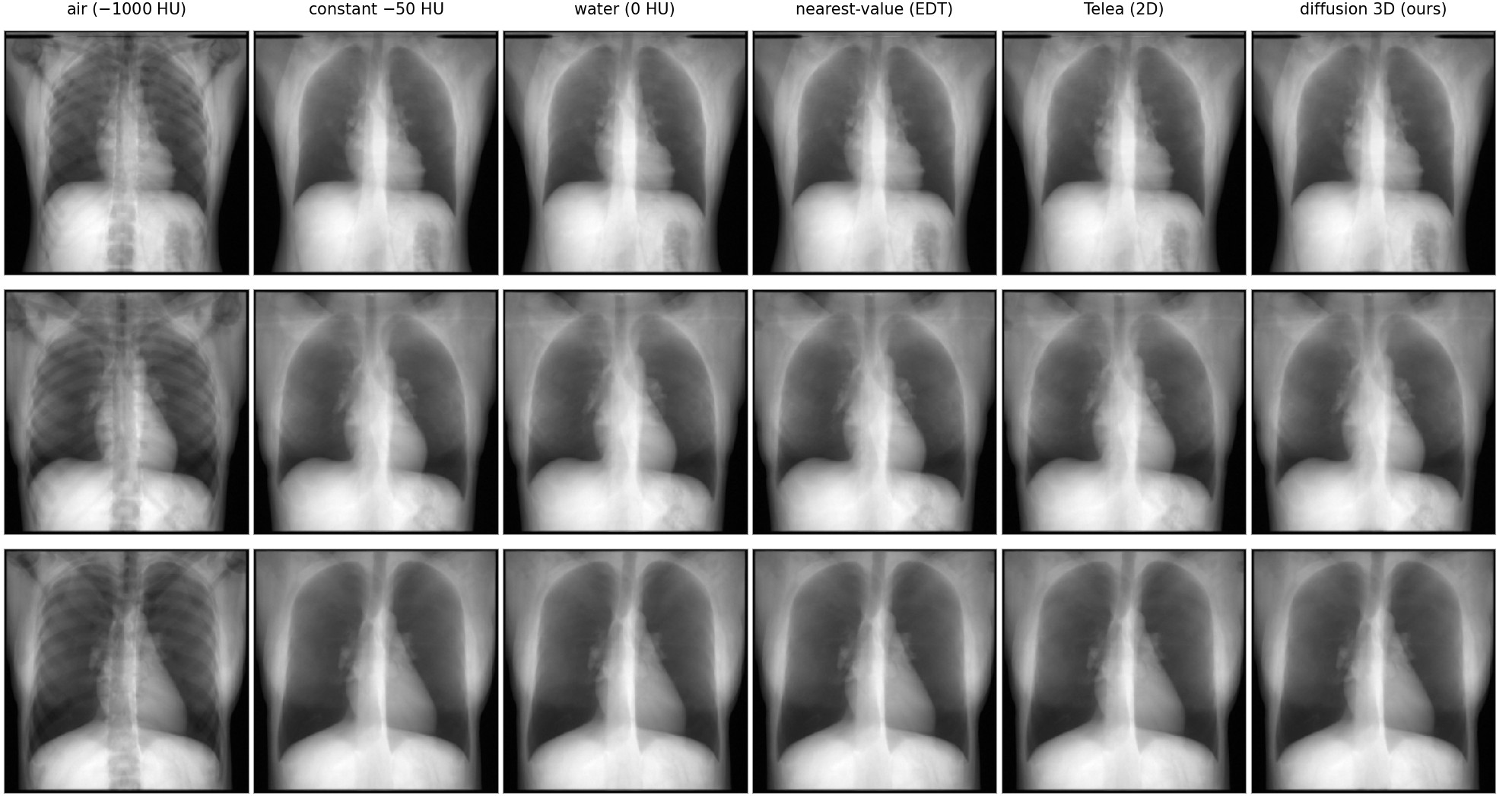}
\put(1.00,53.82){\sffamily\bfseries\small a}
\put(17.67,53.82){\sffamily\bfseries\small b}
\put(34.33,53.82){\sffamily\bfseries\small c}
\put(51.00,53.82){\sffamily\bfseries\small d}
\put(67.67,53.82){\sffamily\bfseries\small e}
\put(84.33,53.82){\sffamily\bfseries\small f}
\end{overpic}
\caption{\textbf{Comparison of soft-tissue void fills.} Panels a--f correspond to the six fills in the order described below. Soft-tissue projections after bone removal, under the six void fills of
the main text's inpainting ablation (columns), for three example CTs (rows). Left
to right: air ($-1000$~HU), constant $-50$~HU, water ($0$~HU), nearest-value
(Euclidean distance transform), slice-wise Telea, and our iterative 3D diffusion
inpainting. All six fills are applied to the same effective-density field (after
the HU-to-density transfer function), so the comparison changes only the fill rule,
not the domain; the constant fills are expressed here by their HU value for
readability. The diffusion fill is Algorithm~\mainref{alg:inpaint}{2} of the main manuscript. The air fill leaves rib-shaped imprints over
the parenchyma; the diffusion fill leaves the least bone-shaped residue.}
\label{supp-fig:inpaint_qual}
\end{figure}

Figure~\ref{supp-fig:inpaint_qual} shows the qualitative counterpart of the void-fill
ablation in the main manuscript (Section~\mainref{sec:ablation}{2.5}): the soft-tissue projection of three example
CTs rendered after bone removal, under each of the six fills. The air fill leaves
a strong rib-shaped imprint across the whole thorax; the remaining fills are close
at this scale, consistent with the small quantitative margins of the main-text
table, with the diffusion fill leaving the least bone-shaped residue.

\section{Bone removal versus detail retention on a common 256 by 256 grid}
\label{supp-sec:supp_antiblur256}
Table~\ref{supp-tab:supp_antiblur256} repeats the main removal/retention analysis with \emph{every} method, input and output, resampled to the $256^2$ native grid of CXR-BS, the band-pass and dilation rescaled accordingly (rib width $3$~px). This is the apples-to-apples comparison for CXR-BS; for the $1024^2$-native methods it discards the fine detail on which main Table~\mainref{tab:antiblur}{6} is scored, so the differences among them are compressed. The removal ordering is unchanged: the diffusion methods remove more rib-band energy than ours on every dataset. Retention differs between the two grids: at $1024^2$ the diffusion methods' retention is below ours on Node21 but exceeds unity, and ours, on VinDr and JSRT ($1.23$--$1.61$ versus $1.05$--$1.19$), whereas at $256^2$ it is below ours on every dataset ($0.33$--$0.76$ versus $0.88$--$0.92$).

\section{Additional illustrations}
\begin{figure}[H]
\centering
\vspace{6mm}
\includegraphics[width=160mm]{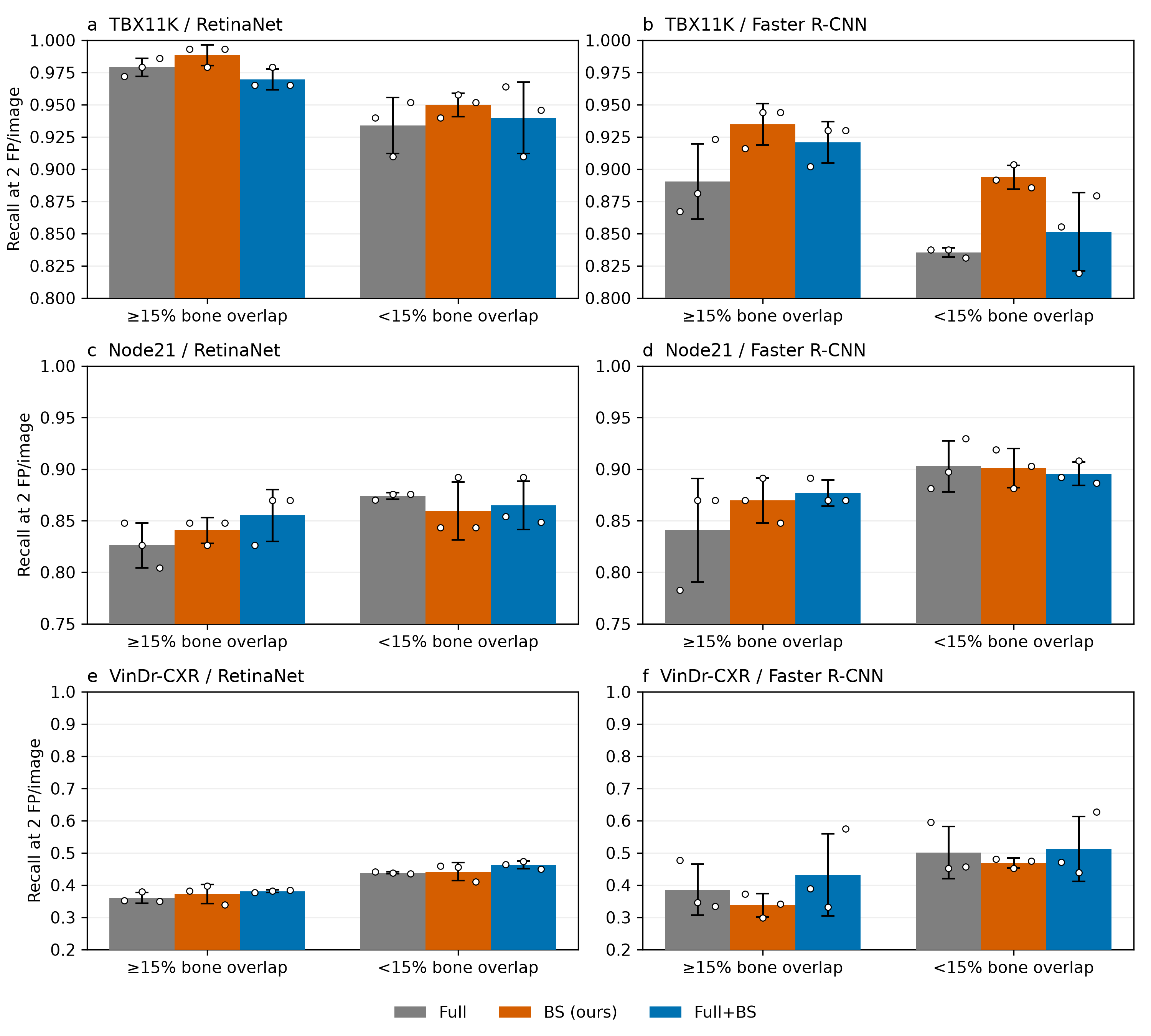}
\caption{\textbf{Recall by bone-overlap group for our suppression method.} Panels a--f show TBX11K, Node21 and VinDr-CXR (rows), with RetinaNet and Faster R-CNN (columns). Recall at 2 false positives per image for lesions with $\ge15\%$ and $<15\%$ bone overlap, plotting main Table~\mainref{tab:subgroup}{2}. Bars show the mean and error bars the sample SD across three training seeds ($n=3$); open circles show the individual seeds. Vertical axes are truncated and differ between datasets.}
\label{supp-fig:subgroup}
\end{figure}

\begin{figure}[H]
\centering
\vspace{6mm}
\begin{overpic}[width=160.0mm]{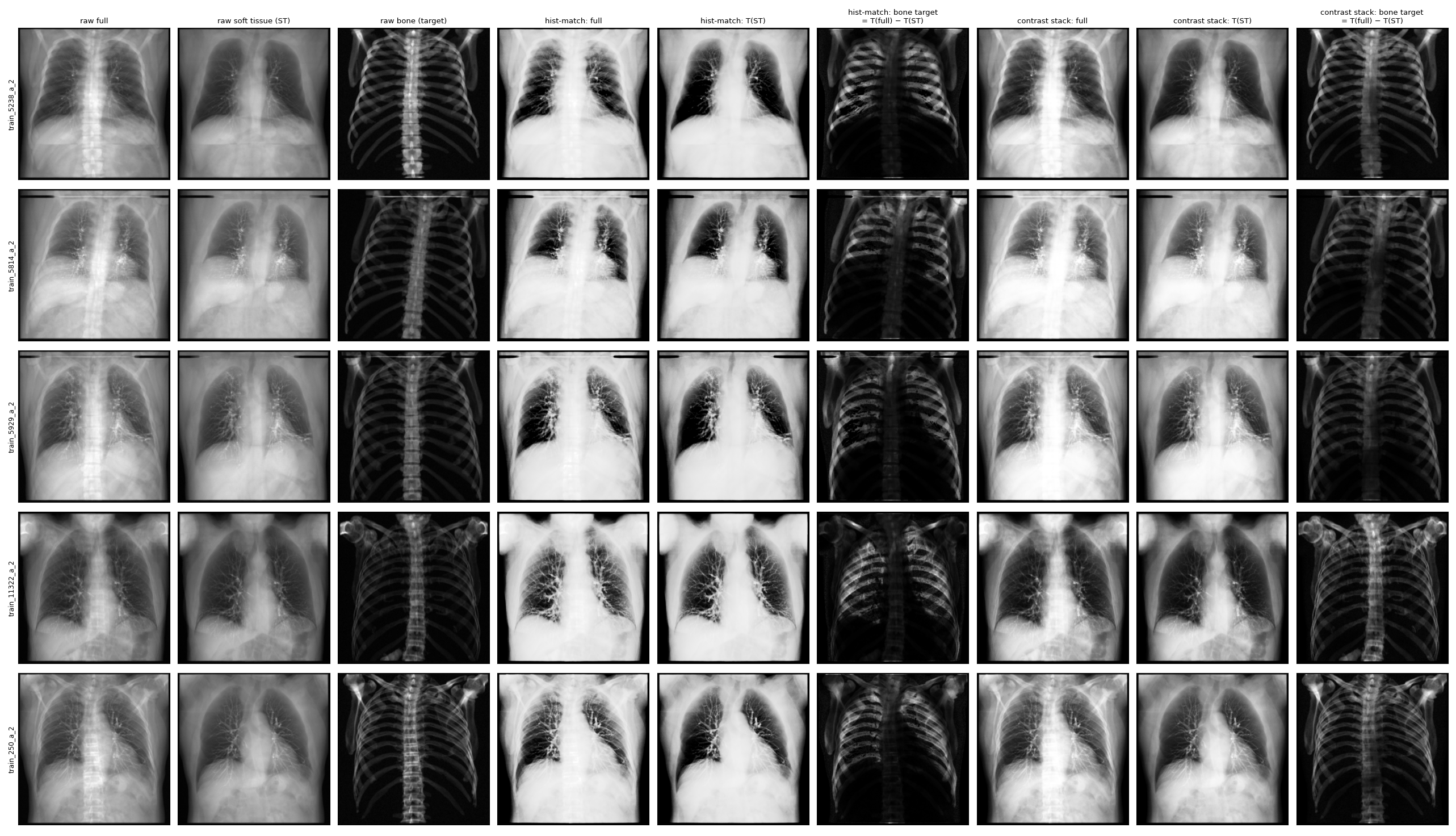}
\put(1.00,58.59){\sffamily\bfseries\small a}
\put(34.33,58.59){\sffamily\bfseries\small b}
\put(67.67,58.59){\sffamily\bfseries\small c}
\end{overpic}
\caption{\textbf{Additivity under appearance augmentation.} Panels a--c group the original, histogram-matched and contrast-augmented triplets, respectively. Five CT-RATE examples illustrate the full image, soft-tissue image and bone target before augmentation and after histogram matching or contrast augmentation. The displayed triplets satisfy the additive identity. Bone panels are individually normalised for display. All augmentation procedures and target recomputation are described in the main Methods.}
\label{supp-fig:supp_aug}
\end{figure}

Supplementary Fig.~\ref{supp-fig:subgroup} visualises the overlap-specific recall values reported in the main manuscript. Supplementary Fig.~\ref{supp-fig:supp_aug} illustrates the additive identity under the appearance augmentations whose complete implementation is described in the main Methods.

\begin{table}[H]
\centering
\caption{BS-Diff, FROC CPM (mean$\pm$SD over seeds 42/43/44) under the main-text
detection protocol. Corresponding \emph{full}-arm baselines (main manuscript, Table~\mainref{tab:detection}{1}):
TBX11K $0.929$/$0.860$, Node21 $0.818$/$0.821$ (RetinaNet/Faster R-CNN).}
\label{supp-tab:bsdiff}
\small
\begin{tabular}{llcc}
\toprule
Dataset & Arm & RetinaNet & Faster R-CNN \\
\midrule
\multirow{2}{*}{TBX11K} & $bs$ & $0.831{\pm}.004$ & $0.741{\pm}.080$ \\
 & $full_{bs}$ & $0.913{\pm}.017$ & $0.876{\pm}.009$ \\
\midrule
\multirow{2}{*}{Node21} & $bs$ & $0.127{\pm}.044$ & $0.159{\pm}.068$ \\
 & $full_{bs}$ & $0.807{\pm}.022$ & $0.813{\pm}.005$ \\
\bottomrule
\end{tabular}
\end{table}
\begin{table}[H]
\centering
\caption{Bone-removal completeness vs.\ clear-lung detail retention with all methods evaluated at $256^2$ (mean$\pm$SD; paired Wilcoxon vs.\ ours, with signed-rank statistics and numerical $p$ values in Supplementary Table~\ref{supp-tab:wilcoxon256}). Retention target $\approx 1$; \textbf{bold} $=$ retention closest to $1$ per dataset (removal left unbolded, as in the main text). $n{=}350$ (Node21), $n{=}3000$ (VinDr), $n{=}247$ (JSRT).}
\label{supp-tab:supp_antiblur256}
\small
\setlength{\tabcolsep}{3pt}
\begin{adjustbox}{max width=\linewidth}
\begin{tabular}{lccccccc}
\toprule
 & Native & \multicolumn{2}{c}{Node21} & \multicolumn{2}{c}{VinDr} & \multicolumn{2}{c}{JSRT} \\
\cmidrule(lr){3-4}\cmidrule(lr){5-6}\cmidrule(lr){7-8}
Method & res & removal & retention & removal & retention & removal & retention \\
\midrule
\textbf{Ours} & $1024$ & $0.31{\pm}.08$ & $\mathbf{0.92{\pm}.22}$ & $0.34{\pm}.08$ & $0.90{\pm}.20$ & $0.25{\pm}.05$ & $\mathbf{0.88{\pm}.10}$ \\
CXR-BS        & $256$  & $0.32{\pm}.03$ & $0.84{\pm}.16$ & $0.31{\pm}.06$ & $\mathbf{0.91{\pm}.20}$ & $0.16{\pm}.03$ & $1.24{\pm}.07$ \\
GL-LCM        & $1024$ & $0.58{\pm}.04$ & $0.34{\pm}.11$ & $0.55{\pm}.06$ & $0.47{\pm}.21$ & $0.43{\pm}.05$ & $0.76{\pm}.23$ \\
DeBoneDiT     & $1024$ & $0.34{\pm}.06$ & $0.41{\pm}.09$ & $0.47{\pm}.08$ & $0.33{\pm}.12$ & $0.40{\pm}.05$ & $0.41{\pm}.08$ \\
BS-LDM        & $1024$ & $0.53{\pm}.08$ & $0.42{\pm}.15$ & $0.47{\pm}.12$ & $0.69{\pm}.39$ & $0.41{\pm}.05$ & $0.72{\pm}.18$ \\
\bottomrule
\end{tabular}
\end{adjustbox}
\end{table}
\clearpage

\clearpage
\begin{table}[H]
\centering
\caption{\textbf{Paired Wilcoxon signed-rank tests at $1024\times1024$ resolution.} Comparisons of our method with each $1024\times1024$-native method for bone removal and detail retention, corresponding to the upper block of main Table~\mainref{tab:antiblur}{6}. $n$ is the number of paired radiographs. $W$ is the smaller of the positive and negative rank sums for the differences (ours minus the listed method). Tests are two-sided, use the normal approximation with tie correction, omit zero differences and apply no continuity correction. Numerical $p$ values are unadjusted for multiple comparisons. These tests assess paired measurement differences; they do not by themselves establish that a method is better, because detail retention targets $1$ and excessive removal can reflect lost image detail.}
\label{supp-tab:wilcoxon1024}
\footnotesize
\setlength{\tabcolsep}{4pt}
\begin{adjustbox}{max width=\linewidth}
\begin{tabular}{llrrrrr}
\toprule
Dataset & Method & $n$ & $W_{\rm removal}$ & $p_{\rm removal}$ & $W_{\rm retention}$ & $p_{\rm retention}$ \\
\midrule
Node21 & GL-LCM & 350 & $0$ & $4.05\times10^{-59}$ & $58$ & $6.67\times10^{-59}$ \\
Node21 & DeBoneDiT & 350 & $20{,}371$ & $4.78\times10^{-8}$ & $263$ & $3.85\times10^{-58}$ \\
Node21 & BS-LDM & 350 & $385$ & $1.09\times10^{-57}$ & $15$ & $4.61\times10^{-59}$ \\
\midrule
VinDr-CXR & GL-LCM & 3000 & $206$ & $4.49\times10^{-491}$ & $2{,}182{,}363$ & $0.1495$ \\
VinDr-CXR & DeBoneDiT & 3000 & $146{,}371$ & $1.21\times10^{-429}$ & $1{,}975{,}297$ & $6.41\times10^{-9}$ \\
VinDr-CXR & BS-LDM & 3000 & $270{,}767$ & $1.32\times10^{-380}$ & $1{,}918{,}072$ & $2.35\times10^{-12}$ \\
\midrule
JSRT & GL-LCM & 247 & $0$ & $2.87\times10^{-42}$ & $822$ & $4.92\times10^{-38}$ \\
JSRT & DeBoneDiT & 247 & $0$ & $2.87\times10^{-42}$ & $6{,}435$ & $2.8\times10^{-15}$ \\
JSRT & BS-LDM & 247 & $0$ & $2.87\times10^{-42}$ & $1{,}472$ & $7.54\times10^{-35}$ \\
\bottomrule
\end{tabular}
\end{adjustbox}
\end{table}

\clearpage
\begin{table}[H]
\centering
\caption{\textbf{Paired Wilcoxon signed-rank tests at $256\times256$ resolution.} Comparisons of our method with each listed method for bone removal and detail retention with all outputs resampled to $256\times256$, corresponding to Supplementary Table~\ref{supp-tab:supp_antiblur256}; the CXR-BS rows also correspond to the lower block of main Table~\mainref{tab:antiblur}{6}. $n$ is the number of paired radiographs. $W$ is the smaller of the positive and negative rank sums for the differences (ours minus the listed method). Tests are two-sided, use the normal approximation with tie correction, omit zero differences and apply no continuity correction. Numerical $p$ values are unadjusted for multiple comparisons. These tests assess paired measurement differences; they do not by themselves establish that a method is better, because detail retention targets $1$ and excessive removal can reflect lost image detail.}
\label{supp-tab:wilcoxon256}
\footnotesize
\setlength{\tabcolsep}{4pt}
\begin{adjustbox}{max width=\linewidth}
\begin{tabular}{llrrrrr}
\toprule
Dataset & Method & $n$ & $W_{\rm removal}$ & $p_{\rm removal}$ & $W_{\rm retention}$ & $p_{\rm retention}$ \\
\midrule
Node21 & GL-LCM & 350 & $0$ & $4.05\times10^{-59}$ & $0$ & $4.05\times10^{-59}$ \\
Node21 & DeBoneDiT & 350 & $18{,}369$ & $7.21\times10^{-11}$ & $1$ & $4.09\times10^{-59}$ \\
Node21 & BS-LDM & 350 & $325$ & $6.52\times10^{-58}$ & $88$ & $8.63\times10^{-59}$ \\
Node21 & CXR-BS & 350 & $24{,}384$ & $8.35\times10^{-4}$ & $17{,}792$ & $9.05\times10^{-12}$ \\
\midrule
VinDr-CXR & GL-LCM & 3000 & $379$ & $1.65\times10^{-490}$ & $80{,}745$ & $3.41\times10^{-456}$ \\
VinDr-CXR & DeBoneDiT & 3000 & $143{,}504$ & $2.9\times10^{-430}$ & $525$ & $1.91\times10^{-490}$ \\
VinDr-CXR & BS-LDM & 3000 & $300{,}785$ & $1.26\times10^{-368}$ & $948{,}762$ & $3.84\times10^{-165}$ \\
VinDr-CXR & CXR-BS & 3000 & $1{,}259{,}080$ & $1.98\times10^{-96}$ & $2{,}146{,}999$ & $0.0362$ \\
\midrule
JSRT & GL-LCM & 247 & $0$ & $2.87\times10^{-42}$ & $2{,}616$ & $1.36\times10^{-29}$ \\
JSRT & DeBoneDiT & 247 & $0$ & $2.87\times10^{-42}$ & $0$ & $2.87\times10^{-42}$ \\
JSRT & BS-LDM & 247 & $0$ & $2.87\times10^{-42}$ & $2{,}969$ & $4.61\times10^{-28}$ \\
JSRT & CXR-BS & 247 & $2$ & $2.94\times10^{-42}$ & $0$ & $2.87\times10^{-42}$ \\
\bottomrule
\end{tabular}
\end{adjustbox}
\end{table}

\clearpage
\section{Bone-overlap recall across suppression methods}
All methods use the same lesion groups as main Table~\mainref{tab:subgroup}{2}. Recall is evaluated at 2 false positives per image, with bone-map intensity $>0.12$ and lesion-box overlap $\ge15\%$. Values are means $\pm$ sample SD across seeds 42, 43 and 44.
\begin{table}[H]
\centering
\caption{TBX11K recall by bone-overlap group across suppression methods. $\Delta$ is the mean seed-matched change from the original-image baseline. Bold marks the highest mean recall within each suppression-input block and detector; rankings are descriptive.}\label{supp-tab:overlap_tbx11k}
\textbf{A. RetinaNet}\par\medskip
\small\begin{adjustbox}{max width=\linewidth}\begin{tabular}{llcccccc}\toprule Input&Method&Seeds&Overlap recall&$\Delta_{ov}$&Clearer recall&$\Delta_{clear}$&$\Delta_{ov}-\Delta_{clear}$\\\midrule
Original & -- & 42/43/44 & $0.979\pm0.007$ & -- & $0.934\pm0.022$ & -- & --\\
Suppressed & Ours & 42/43/44 & $\mathbf{0.988\pm0.008}$ & $+0.009$ & $\mathbf{0.950\pm0.009}$ & $+0.016$ & $-0.007$\\
Suppressed & CXR-BS & 42/43/44 & $0.951\pm0.019$ & $-0.028$ & $0.932\pm0.018$ & $-0.002$ & $-0.026$\\
Suppressed & GL-LCM & 42/43/44 & $0.977\pm0.008$ & $-0.002$ & $0.932\pm0.009$ & $-0.002$ & $-0.000$\\
Suppressed & DeBoneDiT & 42/43/44 & $0.967\pm0.011$ & $-0.012$ & $0.906\pm0.019$ & $-0.028$ & $+0.016$\\
Suppressed & BS-LDM & 42/43/44 & $0.970\pm0.011$ & $-0.009$ & $0.898\pm0.006$ & $-0.036$ & $+0.027$\\
Combined & Ours & 42/43/44 & $0.970\pm0.008$ & $-0.009$ & $\mathbf{0.940\pm0.028}$ & $+0.006$ & $-0.015$\\
Combined & CXR-BS & 42/43/44 & $\mathbf{0.974\pm0.004}$ & $-0.005$ & $\mathbf{0.940\pm0.016}$ & $+0.006$ & $-0.011$\\
Combined & GL-LCM & 42/43/44 & $0.963\pm0.022$ & $-0.016$ & $0.916\pm0.006$ & $-0.018$ & $+0.002$\\
Combined & DeBoneDiT & 42/43/44 & $0.963\pm0.011$ & $-0.016$ & $0.908\pm0.013$ & $-0.026$ & $+0.010$\\
Combined & BS-LDM & 42/43/44 & $0.965\pm0.007$ & $-0.014$ & $0.920\pm0.034$ & $-0.014$ & $+0.000$\\
\bottomrule\end{tabular}\end{adjustbox}
\par\bigskip
\textbf{B. Faster R-CNN}\par\medskip
\small\begin{adjustbox}{max width=\linewidth}\begin{tabular}{llcccccc}\toprule Input&Method&Seeds&Overlap recall&$\Delta_{ov}$&Clearer recall&$\Delta_{clear}$&$\Delta_{ov}-\Delta_{clear}$\\\midrule
Original & -- & 42/43/44 & $0.890\pm0.029$ & -- & $0.835\pm0.003$ & -- & --\\
Suppressed & Ours & 42/43/44 & $\mathbf{0.935\pm0.016}$ & $+0.044$ & $\mathbf{0.894\pm0.009}$ & $+0.058$ & $-0.014$\\
Suppressed & CXR-BS & 42/43/44 & $0.932\pm0.022$ & $+0.042$ & $0.886\pm0.032$ & $+0.050$ & $-0.008$\\
Suppressed & GL-LCM & 42/43/44 & $0.932\pm0.029$ & $+0.042$ & $0.867\pm0.054$ & $+0.032$ & $+0.010$\\
Suppressed & DeBoneDiT & 42/43/44 & $0.921\pm0.018$ & $+0.030$ & $0.867\pm0.022$ & $+0.032$ & $-0.002$\\
Suppressed & BS-LDM & 42/43/44 & $0.881\pm0.074$ & $-0.009$ & $0.825\pm0.079$ & $-0.010$ & $+0.001$\\
Combined & Ours & 42/43/44 & $0.921\pm0.016$ & $+0.030$ & $0.851\pm0.030$ & $+0.016$ & $+0.014$\\
Combined & CXR-BS & 42/43/44 & $0.925\pm0.039$ & $+0.035$ & $0.839\pm0.024$ & $+0.004$ & $+0.031$\\
Combined & GL-LCM & 42/43/44 & $\mathbf{0.935\pm0.035}$ & $+0.044$ & $\mathbf{0.873\pm0.040}$ & $+0.038$ & $+0.006$\\
Combined & DeBoneDiT & 42/43/44 & $0.909\pm0.028$ & $+0.019$ & $0.861\pm0.016$ & $+0.026$ & $-0.007$\\
Combined & BS-LDM & 42/43/44 & $0.902\pm0.050$ & $+0.012$ & $0.839\pm0.057$ & $+0.004$ & $+0.008$\\
\bottomrule\end{tabular}\end{adjustbox}
\par\bigskip
\end{table}
\begin{table}[H]
\centering
\caption{Node21 recall by bone-overlap group across suppression methods. $\Delta$ is the mean seed-matched change from the original-image baseline. Bold marks the highest mean recall within each suppression-input block and detector; rankings are descriptive.}\label{supp-tab:overlap_node21}
\textbf{A. RetinaNet}\par\medskip
\small\begin{adjustbox}{max width=\linewidth}\begin{tabular}{llcccccc}\toprule Input&Method&Seeds&Overlap recall&$\Delta_{ov}$&Clearer recall&$\Delta_{clear}$&$\Delta_{ov}-\Delta_{clear}$\\\midrule
Original & -- & 42/43/44 & $0.826\pm0.022$ & -- & $0.874\pm0.003$ & -- & --\\
Suppressed & Ours & 42/43/44 & $\mathbf{0.841\pm0.013}$ & $+0.014$ & $\mathbf{0.859\pm0.028}$ & $-0.014$ & $+0.029$\\
Suppressed & CXR-BS & 42/43/44 & $0.725\pm0.070$ & $-0.101$ & $0.784\pm0.005$ & $-0.090$ & $-0.011$\\
Suppressed & GL-LCM & 42/43/44 & $0.623\pm0.091$ & $-0.203$ & $0.650\pm0.023$ & $-0.223$ & $+0.021$\\
Suppressed & DeBoneDiT & 42/43/44 & $0.688\pm0.055$ & $-0.138$ & $0.813\pm0.017$ & $-0.061$ & $-0.076$\\
Suppressed & BS-LDM & 42/43/44 & $0.399\pm0.066$ & $-0.428$ & $0.586\pm0.045$ & $-0.288$ & $-0.139$\\
Combined & Ours & 42/43/44 & $\mathbf{0.855\pm0.025}$ & $+0.029$ & $0.865\pm0.024$ & $-0.009$ & $+0.038$\\
Combined & CXR-BS & 42/43/44 & $0.783\pm0.043$ & $-0.043$ & $0.850\pm0.031$ & $-0.023$ & $-0.020$\\
Combined & GL-LCM & 42/43/44 & $0.819\pm0.013$ & $-0.007$ & $0.852\pm0.052$ & $-0.022$ & $+0.014$\\
Combined & DeBoneDiT & 42/43/44 & $0.761\pm0.043$ & $-0.065$ & $0.858\pm0.017$ & $-0.016$ & $-0.049$\\
Combined & BS-LDM & 42/43/44 & $0.768\pm0.013$ & $-0.058$ & $\mathbf{0.876\pm0.014}$ & $+0.002$ & $-0.060$\\
\bottomrule\end{tabular}\end{adjustbox}
\par\bigskip
\textbf{B. Faster R-CNN}\par\medskip
\small\begin{adjustbox}{max width=\linewidth}\begin{tabular}{llcccccc}\toprule Input&Method&Seeds&Overlap recall&$\Delta_{ov}$&Clearer recall&$\Delta_{clear}$&$\Delta_{ov}-\Delta_{clear}$\\\midrule
Original & -- & 42/43/44 & $0.841\pm0.050$ & -- & $0.903\pm0.025$ & -- & --\\
Suppressed & Ours & 42/43/44 & $\mathbf{0.870\pm0.022}$ & $+0.029$ & $\mathbf{0.901\pm0.019}$ & $-0.002$ & $+0.031$\\
Suppressed & CXR-BS & 42/43/44 & $0.710\pm0.045$ & $-0.130$ & $0.798\pm0.017$ & $-0.105$ & $-0.026$\\
Suppressed & GL-LCM & 42/43/44 & $0.659\pm0.055$ & $-0.181$ & $0.728\pm0.042$ & $-0.175$ & $-0.006$\\
Suppressed & DeBoneDiT & 42/43/44 & $0.659\pm0.076$ & $-0.181$ & $0.818\pm0.006$ & $-0.085$ & $-0.096$\\
Suppressed & BS-LDM & 42/43/44 & $0.609\pm0.022$ & $-0.232$ & $0.645\pm0.022$ & $-0.258$ & $+0.026$\\
Combined & Ours & 42/43/44 & $\mathbf{0.877\pm0.013}$ & $+0.036$ & $\mathbf{0.895\pm0.011}$ & $-0.007$ & $+0.043$\\
Combined & CXR-BS & 42/43/44 & $0.783\pm0.058$ & $-0.058$ & $\mathbf{0.895\pm0.008}$ & $-0.007$ & $-0.051$\\
Combined & GL-LCM & 42/43/44 & $0.848\pm0.058$ & $+0.007$ & $0.888\pm0.012$ & $-0.014$ & $+0.022$\\
Combined & DeBoneDiT & 42/43/44 & $0.826\pm0.022$ & $-0.014$ & $0.874\pm0.019$ & $-0.029$ & $+0.014$\\
Combined & BS-LDM & 42/43/44 & $0.804\pm0.065$ & $-0.036$ & $0.876\pm0.044$ & $-0.027$ & $-0.009$\\
\bottomrule\end{tabular}\end{adjustbox}
\par\bigskip
\end{table}
\begin{table}[H]
\centering
\caption{VinDr-CXR recall by bone-overlap group across suppression methods. $\Delta$ is the mean seed-matched change from the original-image baseline. Bold marks the highest mean recall within each suppression-input block and detector; rankings are descriptive.}\label{supp-tab:overlap_vindr}
\textbf{A. RetinaNet}\par\medskip
\small\begin{adjustbox}{max width=\linewidth}\begin{tabular}{llcccccc}\toprule Input&Method&Seeds&Overlap recall&$\Delta_{ov}$&Clearer recall&$\Delta_{clear}$&$\Delta_{ov}-\Delta_{clear}$\\\midrule
Original & -- & 42/43/44 & $0.360\pm0.017$ & -- & $0.438\pm0.003$ & -- & --\\
Suppressed & Ours & 42/43/44 & $\mathbf{0.373\pm0.030}$ & $+0.013$ & $\mathbf{0.442\pm0.028}$ & $+0.004$ & $+0.009$\\
Suppressed & CXR-BS & 42/43/44 & $0.326\pm0.016$ & $-0.034$ & $0.402\pm0.027$ & $-0.036$ & $+0.002$\\
Suppressed & GL-LCM & 42/43/44 & $0.324\pm0.019$ & $-0.036$ & $0.390\pm0.014$ & $-0.048$ & $+0.012$\\
Suppressed & DeBoneDiT & 42/43/44 & $0.351\pm0.009$ & $-0.009$ & $0.401\pm0.038$ & $-0.037$ & $+0.028$\\
Suppressed & BS-LDM & 42/43/44 & $0.307\pm0.032$ & $-0.054$ & $0.354\pm0.031$ & $-0.084$ & $+0.030$\\
Combined & Ours & 42/43/44 & $0.381\pm0.004$ & $+0.021$ & $\mathbf{0.463\pm0.012}$ & $+0.024$ & $-0.003$\\
Combined & CXR-BS & 42/43/44 & $0.377\pm0.020$ & $+0.017$ & $0.433\pm0.023$ & $-0.006$ & $+0.022$\\
Combined & GL-LCM & 42/43/44 & $\mathbf{0.389\pm0.029}$ & $+0.028$ & $0.446\pm0.012$ & $+0.008$ & $+0.021$\\
Combined & DeBoneDiT & 42/43/44 & $0.360\pm0.019$ & $+0.000$ & $0.400\pm0.033$ & $-0.038$ & $+0.038$\\
Combined & BS-LDM & 42/43/44 & $0.363\pm0.026$ & $+0.003$ & $0.418\pm0.010$ & $-0.020$ & $+0.024$\\
\bottomrule\end{tabular}\end{adjustbox}
\par\bigskip
\textbf{B. Faster R-CNN}\par\medskip
\small\begin{adjustbox}{max width=\linewidth}\begin{tabular}{llcccccc}\toprule Input&Method&Seeds&Overlap recall&$\Delta_{ov}$&Clearer recall&$\Delta_{clear}$&$\Delta_{ov}-\Delta_{clear}$\\\midrule
Original & -- & 42/43/44 & $0.386\pm0.079$ & -- & $0.501\pm0.081$ & -- & --\\
Suppressed & Ours & 42/43/44 & $0.338\pm0.037$ & $-0.049$ & $0.469\pm0.015$ & $-0.032$ & $-0.016$\\
Suppressed & CXR-BS & 42/43/44 & $0.254\pm0.025$ & $-0.132$ & $0.384\pm0.016$ & $-0.118$ & $-0.015$\\
Suppressed & GL-LCM & 42/43/44 & $0.343\pm0.086$ & $-0.044$ & $0.460\pm0.061$ & $-0.042$ & $-0.002$\\
Suppressed & DeBoneDiT & 42/43/44 & $0.287\pm0.024$ & $-0.099$ & $0.426\pm0.035$ & $-0.075$ & $-0.024$\\
Suppressed & BS-LDM & 42/43/44 & $\mathbf{0.405\pm0.039}$ & $+0.018$ & $\mathbf{0.511\pm0.047}$ & $+0.010$ & $+0.009$\\
Combined & Ours & 42/43/44 & $\mathbf{0.432\pm0.127}$ & $+0.046$ & $\mathbf{0.512\pm0.101}$ & $+0.011$ & $+0.035$\\
Combined & CXR-BS & 42/43/44 & $0.392\pm0.047$ & $+0.006$ & $0.503\pm0.046$ & $+0.002$ & $+0.004$\\
Combined & GL-LCM & 42/43/44 & $0.376\pm0.042$ & $-0.010$ & $0.475\pm0.046$ & $-0.026$ & $+0.016$\\
Combined & DeBoneDiT & 42/43/44 & $0.363\pm0.101$ & $-0.023$ & $0.475\pm0.107$ & $-0.026$ & $+0.003$\\
Combined & BS-LDM & 42/43/44 & $0.357\pm0.105$ & $-0.029$ & $0.466\pm0.093$ & $-0.035$ & $+0.006$\\
\bottomrule\end{tabular}\end{adjustbox}
\par\bigskip
\end{table}

\clearpage
\section{Class-level bone-overlap prevalence in VinDr-CXR}
Supplementary Table~\ref{supp-tab:class_prevalence} lists the class-level values underlying the correlations in main Table~\mainref{tab:perclass}{3} for our method.
\begin{table}[H]
\centering
\caption{Bone-overlap prevalence and detection change for our suppressed-only input across the 14 VinDr-CXR finding classes (official test set; $2{,}632$ findings, $398$ overlapped). A finding is overlapped when at least $15\%$ of its box intersects the predicted bone mask. $\Delta$CPM is the suppressed-minus-original class-level CPM, calculated within each training seed and averaged over seeds 42, 43 and 44. Classes are ordered by prevalence.}
\label{supp-tab:class_prevalence}
\small
\begin{tabular}{lrrrcc}
\toprule
 & & & Prevalence & \multicolumn{2}{c}{$\Delta$CPM ($bs-full$)} \\
\cmidrule(lr){5-6}
Class & Findings & Overlapped & (\%) & RetinaNet & Faster R-CNN \\
\midrule
Pneumothorax & 25 & 9 & 36.0 & $+0.046$ & $+0.002$ \\
Pleural thickening & 240 & 79 & 32.9 & $-0.013$ & $-0.035$ \\
Infiltration & 79 & 25 & 31.6 & $+0.030$ & $+0.009$ \\
Other lesion & 112 & 25 & 22.3 & $+0.030$ & $-0.020$ \\
Pulmonary fibrosis & 358 & 65 & 18.2 & $+0.023$ & $+0.029$ \\
Aortic enlargement & 221 & 37 & 16.7 & $+0.028$ & $-0.015$ \\
Nodule/Mass & 286 & 41 & 14.3 & $+0.023$ & $-0.027$ \\
ILD & 316 & 42 & 13.3 & $-0.002$ & $+0.003$ \\
Calcification & 232 & 29 & 12.5 & $+0.011$ & $+0.012$ \\
Consolidation & 126 & 15 & 11.9 & $+0.010$ & $+0.026$ \\
Lung opacity & 95 & 10 & 10.5 & $+0.006$ & $-0.051$ \\
Atelectasis & 96 & 7 & 7.3 & $+0.049$ & $-0.005$ \\
Pleural effusion & 136 & 9 & 6.6 & $-0.002$ & $-0.065$ \\
Cardiomegaly & 310 & 5 & 1.6 & $-0.052$ & $-0.042$ \\
\bottomrule
\end{tabular}
\end{table}
\end{document}